# A Generative Machine Learning Model for Material Microstructure 3D Reconstruction and Performance Evaluation

Yilin Zheng[1], Zhigong Song[1*]

School of Mechanical Engineering, University of Jiangnan 214122 Wu Xi

song_jnu@jiangnan.edu.cn

**Abstract:** The reconstruction of 3D microstructures from 2D slices is considered to hold significant value in predicting the spatial structure and physical properties of materials. The dimensional extension from 2D to 3D is viewed as a highly challenging inverse problem from the current technological perspective. Recently, methods based on generative adversarial networks have garnered widespread attention. However, they are still hampered by numerous limitations, including oversimplified models, a requirement for a substantial number of training samples, and difficulties in achieving model convergence during training. In light of this, a novel generative model that integrates the multiscale properties of U-net with and the generative capabilities of GAN has been proposed. Based on this, the innovative construction of a multi-scale channel aggregation module, a multi-scale hierarchical feature aggregation module and a convolutional block attention mechanism can better capture the properties of the material microstructure and extract the image information. The model's accuracy is further improved by combining the image regularization loss with the Wasserstein distance loss. In addition, this study utilizes the anisotropy index to accurately distinguish the nature of the image, which can clearly determine the isotropy and anisotropy of the image. It is also the first time that the generation quality of material samples from different domains is evaluated and the performance of the model itself is compared. The experimental results demonstrate that the present model not only shows a very high similarity between the generated 3D structures and real samples but is also highly consistent with real data in terms of statistical data analysis.

## 1. Introduction

Recent years, artificial intelligence generated content (AIGC) [1] has been intensively developed in three-dimensional (3D) reconstruction generation [2]. This method can invert 3D structures from a limited number of two-dimensional (2D) images, which offer unparalleled convenience across various industries. For instance, in the aero domain, reconstructing the arrangement and texture of metallic phases in high-performance alloys [3] contributes to enhancing alloy performance and longevity. In biomedical engineering, the reconstruction of bone constituents such as calcium, collagen and vascular systems [4] provides critical insights for advancements in fracture repair technologies. In soil science, the reconstruction of micro-components' structure [5-6] can improve soil quality and minimize erosion. In the domain of electrochemistry, the reconstruction of electrode porous structures [7-8] can ensure efficient transmission of electrons and ions. Overall, 3D reconstruction techniques based on generative machine learning provide the necessary guarantees for in-depth analysis of material properties and comprehensive understanding of their macroscopic characteristics. Therefore, how to utilize the method of extending from 2D images to 3D structures has become the focus of current research.

Currently, high-precision scanning equipment is relied upon by traditional physical imaging technology to slice and scan 3D structural samples to obtain 2D cross-sections. Subsequently, the 2D cross-sectional images are processed through image synthesis such as stitching and superposition to generate 3D structural models. Such as micro-computerized tomography (micro-CT) [9-10], focused ion beam scanning electron microscopy (FIB-SEM) [11-14], laser scanning confocal microscopy [15-16], serial section tomography (SSTM) [17] and X-ray computed micro-tomography (XCT) are employed. These methods tend to be recognized as inefficient, expensive and time-consuming. Moreover, the samples are often constrained by their inherent imaging mechanisms, which result in their destruction during the imaging process. Consequently, in order to reconstruct the 3D structure of materials more accurately, researchers are increasingly attracted by traditional statistical numerical simulations [18-20] and deep learning methods [21-24].

Traditional numerical simulation approaches rely on 2D images as input and use the statistical characteristics of the materials as constraints for 3D reconstruction. This encompasses approaches such as the simulated annealing (SA) [25-26], Gaussian random field [33-34] and multiple-point statistics (MPS) [38]. Among them, the SA approach [25-26] is capable of being integrated with various cost functions to provide rich morphological and statistical information for reconstruction. Nevertheless, its significant computational demands lead to efficiency issues. To address this challenge, Tang et al [27] have introduced a selection strategy to deal with pixels in different phase neighborhoods, effectively accelerating the convergence rate of the model. AN et al. [28] have advanced this method, achieving enhancement in the model's computational efficiency under the same conditions. Simultaneously, Pant [29] and Jiao et al. [30] have contributed improvements in the realm of statistical functional methods, which improved the efficiency of material microstructure reconstruction. Besides efficiency, the precision of reconstruction is also a key issue to be solved by the simulated annealing method. Addressing this, researchers have employed the two-point correlation function (TPCF) as a means to constrain the higher-order 3D features of microstructures. Since this TPCF constraint is not accurate in constraining higher-order features, Gerke et al. [31] proposed using weighted correlation functions to improve and strengthen these constraints, enhancing the effectiveness and accuracy of reconstruction. Quiblie et al. [32] used the Gaussian random field method [33-34] to reconstruct non-uniform porous materials, but this single constraint method is not enough to reconstruct multi-phase and anisotropic materials. To address this limitation, Jiao et al. [30][35] developed and extended an innovative algorithm for the reconstruction of anisotropic material structures, applying it to the domains of energy and tri-phase lithium-ion battery materials. Suzue et al. [36] applied TPCF for reconstructing microstructures of porous composite battery anodes from 2D phase diagrams. Siddique et al. [37] proposed a stochastic algorithm based on the grain nucleation and growth, aimed at constructing the architecture of 3D fuel cell catalyst layers. Strebelle's team [38] introduced a MPS method of capturing multi-point features by scanning training images using specific templates to reproduce 3D structures pixel by pixel. However, a dramatic increase in memory consumption is observed when large models are used for reconstructing microstructures. To overcome this, a cross-correlation simulation (CCSIM) approach was proposed by Tahmasebi's team [39], which captures the long-range connectivity of porous materials and enables the generation of structures close to real samples through simulation. Despite these advancements, the performance of anisotropic materials in structural reconstruction remains unsatisfactory. Some researchers have proposed improved CCSIM approaches and conducted extended research on the material microstructure from 2D to 3D [40-41].

Accompanied by the significant improvement in computer performance, groundbreaking progress has been achieved in the domain of deep learning. Especially in material structure reconstruction, this progress is largely due to the exceptional data processing and learning prowess exhibited. Distinct from traditional reconstruction techniques, generative machine learning approaches use neural networks as feature extractors, enabling the automated extraction of complex features from images. By the design of a multilevel network structure, not only can the basic features in the image be extracted, but more complex and abstract attributes can also be revealed. The training and reconstruction processes of deep neural network models are separated. Once the model masters the intrinsic properties of the data samples in the training phase, it can quickly generate synthetic samples in the reconstruction phase. These synthetic samples are morphologically highly similar to the original samples, a process that takes only a few seconds. Lee et al. [42] and the Düreth team [43] introduced a denoising diffusion probability model, which extends the reconstruction range from 2D-2D to 2D-3D [16]. This requires the fitting of a diffusion model and estimating an array of critical parameters, such as diffusion rates and acceptance thresholds. Nonetheless, the accurate estimation of these parameters faces significant challenges. Meanwhile, generative adversarial networks (GANs) and their derived models have been widely utilized in the field of material microstructure reconstruction. Mosser et al. [44] used the 3D-DCGAN method to reconstruct the pore structure of porous materials in a solid state. They further extended Mosser et al. method to be applicable to gray-scale image reconstruction [45].

Recently, He et al. [46] applied a recurrent neural network model for sequentially stacking 2D slices to achieve 3D reconstruction. The team used the temporal characteristics in recurrent neural networks (RNN) to effectively resolve the spatial dependencies and continuity problems between adjacent slices. However, when applied to anisotropic materials, the method's limitations become apparent due to insufficient randomness and diversity in spatial dimension expansion. A convolutional neural network featuring a mixed receptive field was designed by Zhang et al. [47], which is able to adapt to feature mapping at different reconstruction stages. Additionally, they introduced a cross-sectional loss function aimed at enhancing the spatial continuity of the structure. He and Zhang [48] further proposed a model PM-ARNN that combines GAN and RNN, which is specifically designed for the use of porous materials in reconstruction. Furthermore, researchers such as He et al. continued to explore the combination of GAN with other neural networks to improve the reconstruction methods for material structures. For example, Zhang and Teng [49] successfully merged GAN with a variational autoencoder (VAE) to reconstruct 3D porous media structures from singular 2D images. Further, Teng [50] reconstructed a large-size 3D model by integrating subtle small-scale information to a macroscopic scale. However, GANs have consistently faced problems associated with gradient vanishing and exploding. To solve these difficulties, researchers have proposed a series of improvements. Gulrajani et al. [51] developed the Wasserstein GAN with Gradient Penalty (WGAN-GP) and Arjovsky [52] advanced the Wasserstein GAN. Zhang and Xia et al. [53-55] implemented modifications to GAN network models and their loss functions. Nevertheless, challenges including the instability of GANs, the requirement for a large number of training samples and high GPU hardware costs have not been completely addressed.

In view of the above limitations, this study aims to solve the shortcomings of existing methods by proposing a new end-to-end generative adversarial network model integrating U-net network architecture [56-57]. The network captures the global morphological features of material structures more acutely. This significantly optimizes the quality of the generated structures, making them closer to the real material structures. Finally, it demonstrates good generalization by reconstructing material microstructures from different domains. The main contributions include:

(1) In this research, a generative model is innovatively proposed, which initially constructs a multi-scale hierarchical feature aggregation network (MHFA). The efficiency and stability of the generative network are improved by integrating multi-scale features. Additionally, a dense hierarchical feature connection strategy is employed to generate features with rich contextual information.

(2) In order to further enhance the local details in the global features, a multi-scale channel

aggregation module (MSCA) is introduced, aiming to enhance the expressiveness of the network and the generalization of the model. Moreover, two different attention mechanisms, spatial and channel, are implemented, which not only enriches the diversity of generated features but also reduces the computational complexity in time and space.

(3) In this study, the GAN theory and the multi-scale concept of U-net are utilized to realize the expansion of material microstructure dimensions from 2D to 3D. The combination of the image regularization loss function and the Wasserstein loss function is introduced to achieve model supervision.

(4) The proposed algorithm model not only demonstrates broad applicability but also deepens the understanding of material properties through a comprehensive evaluation of the materials' generative quality and the model's own performance.

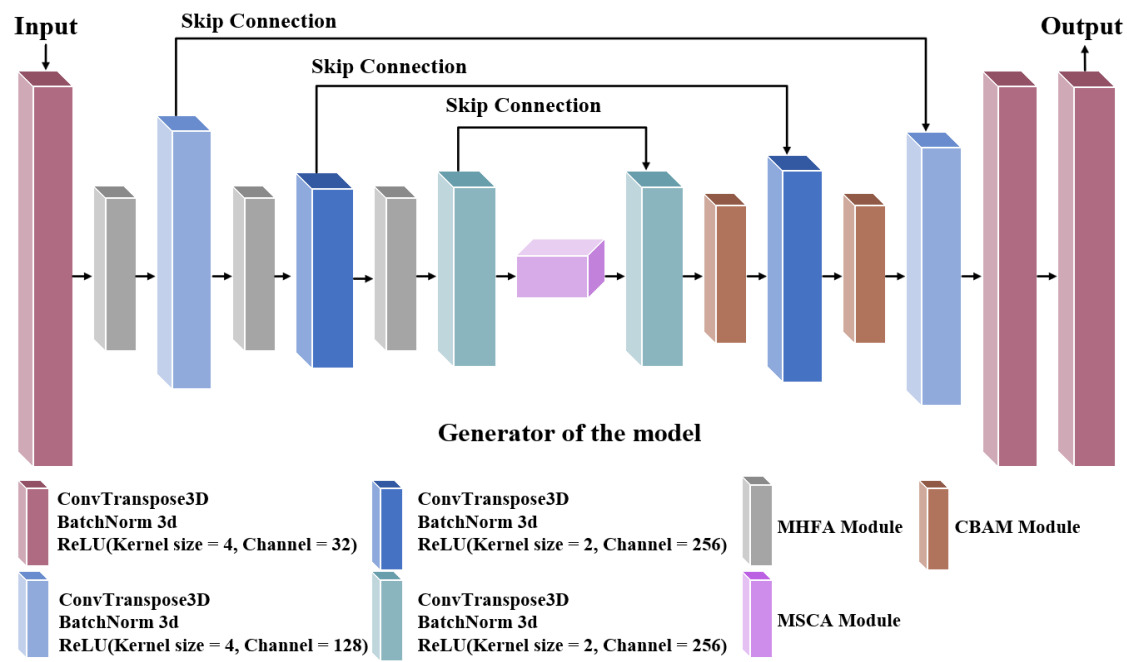

Fig. 1 The proposed generator architecture

## 2 Results and discussion

Fig. 1 depicts the algorithm generator framework in this research. A multi-scale hierarchical feature aggregation network has been integrated to precisely extracting key contextual multi-scale features and semantic information within the microstructures. Additionally, a multi-scale channel aggregation module has been introduced in the intermediate layers of the network. This enhances the model's capability to parse the geometric structure and semantic content of images. Subsequent to the MSCA, a convolutional block attention module (CBAM) is implemented. This optimizes the model's adjustment of channel and spatial weight information, thereby further strengthening the efficacy of feature extraction as shown in Fig 2. A fusion strategy has been employed to more accurately guide the model's learning process. This combines an image regularization loss function with the Wasserstein loss function, optimizing the efficiency of model supervision. In the discriminator network, the study adopts a multi-layer network to emphasize the focus on image feature information to promote the effectiveness of network training. The specific structures and methods can be found in the materials and methods section.

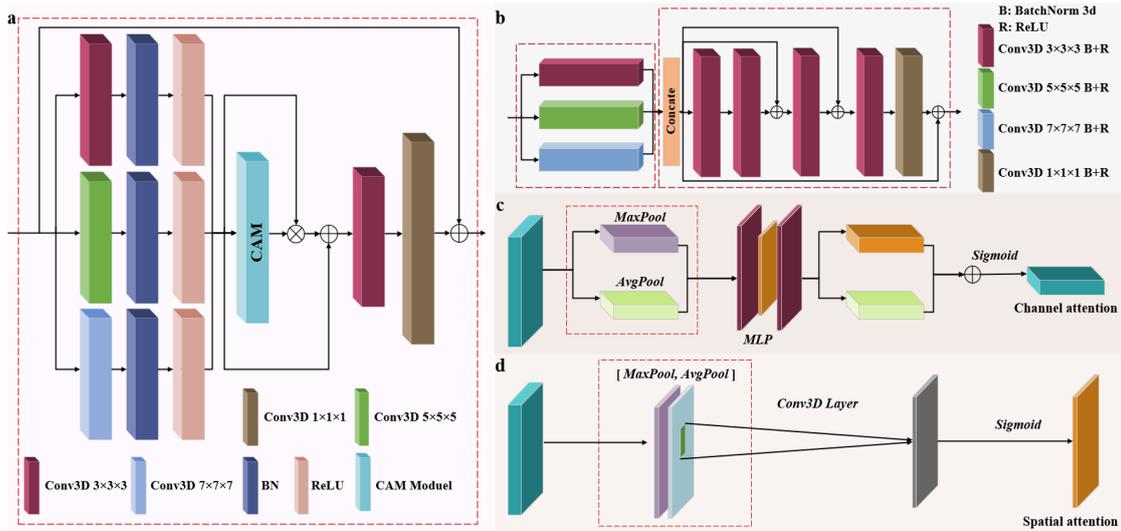

Fig. 2 Different network modules a is the Multi-scale channel aggregation module (MSCA), b is the Multi-scale hierarchical feature aggregation module (MHFA), c is the CAM module, and d is the SAM module (the combination of c and d is known as the CBAM module)

In this section, a comprehensive evaluation is carried out on five selected material samples, focusing on three key aspects: quality assessment of the generated material 3D structures, comparative analysis with current state-of-the-art methods and in-depth discussion of the model performance. This multi-dimensional evaluation approach not only confirms the practicality of the algorithm but also demonstrates its robustness in diverse application scenarios.

## 2.1 Case 1：Berea sandstone

To deeply evaluate the generation quality performance of this paper's algorithm, the Berea Sandstone dataset was initially used to train the samples. As illustrated in Fig. 3, sections a-d, e-h and i-l represent the 3D structure and slices along the X, Y and Z directions of the Sample, Baseline and this paper's model respectively. From these morphology maps, it can be clearly observed that this paper's model performs much better in structural generation compared to Baseline. Further quantitative analysis confirms this by introducing SSIM and PSNR metrics in the image generation domain for comparison. As shown in Table 2, the algorithm model is not only significantly effective in the reconstruction of isotropic porous media but also has higher values of SSIM and PSNR than baseline.

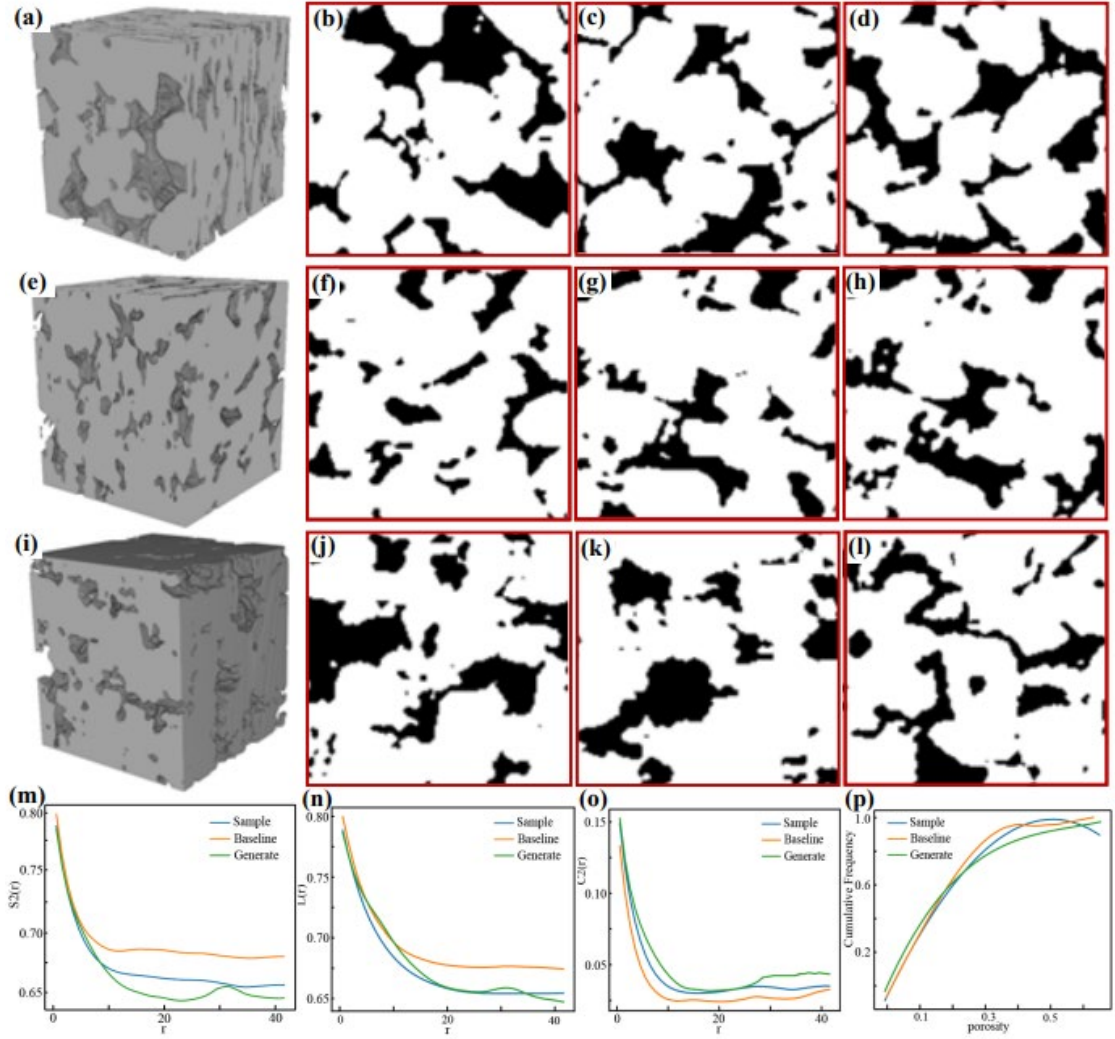

**Fig. 3** 2D and 3D visualizations of representative Berea sandstone samples and experimental statistical results. Where a-d, e-h and i-l are the 3D visualizations of Sample, Baseline and our model, as well as the slices along the different directions of X, Y and Z respectively. Sections m-p are the results of comparison of the average along the different directions S2(r), C2(r), L(r) and the cumulative distribution of local porosity along different directions (real samples in blue, Baseline samples in yellow, and generated samples in green)

  Meanwhile, the performance of the proposed model is evaluated by quantitatively analyzing the experimental results. In Fig. 3 sections m-p demonstrates the average value of each statistical function, indicating that the generated 3D structure has a higher similarity to the real sample in terms of the two-point correlation function. In addition, the generated curves of the linear path function are closer to the real samples compared to baseline. Further, the cumulative distribution plot of local porosity shows that the generated curve converges with the sample curve. This implies that the generated data are highly consistent with the reference data in terms of porosity distribution, thus demonstrating the superior reconstruction effect of our model.

**Table 2** Quantitative comparative metrics of SSIM, PSNR in Berea sandstone samples

| Method | Metrics | |
| --- | --- | --- |
|  | SSIM | PSNR |
| Baseline | 0.462 | 23.496 |
| OUR | **0.575** | **26.900** |

## 2.2 Case 2: Ketton Rock

The Ketton Rock sample [45] [62] is an oolitic limestone, which is utilized for experimental training, and consistency of all parameters is maintained. As shown in Fig. 4 m-p, compared with Baseline (i.e., GAN network), the reconstructed 3D structure is very close to the original sample and displays rich diversity. The result confirms the advantages of this paper's model. Table 3 demonstrates the quantitative comparison of the average SSIM and PSNR of different oriented slice generated by Baseline and our model.

Table 3 Quantitative comparative metrics of SSIM, PSNR in Ketton rock samples

| Method | Metrics | |
|---|---|---|
| | SSIM | PSNR |
| Baseline | 0.208 | 23.506 |
| OUR | **0.293** | **24.321** |

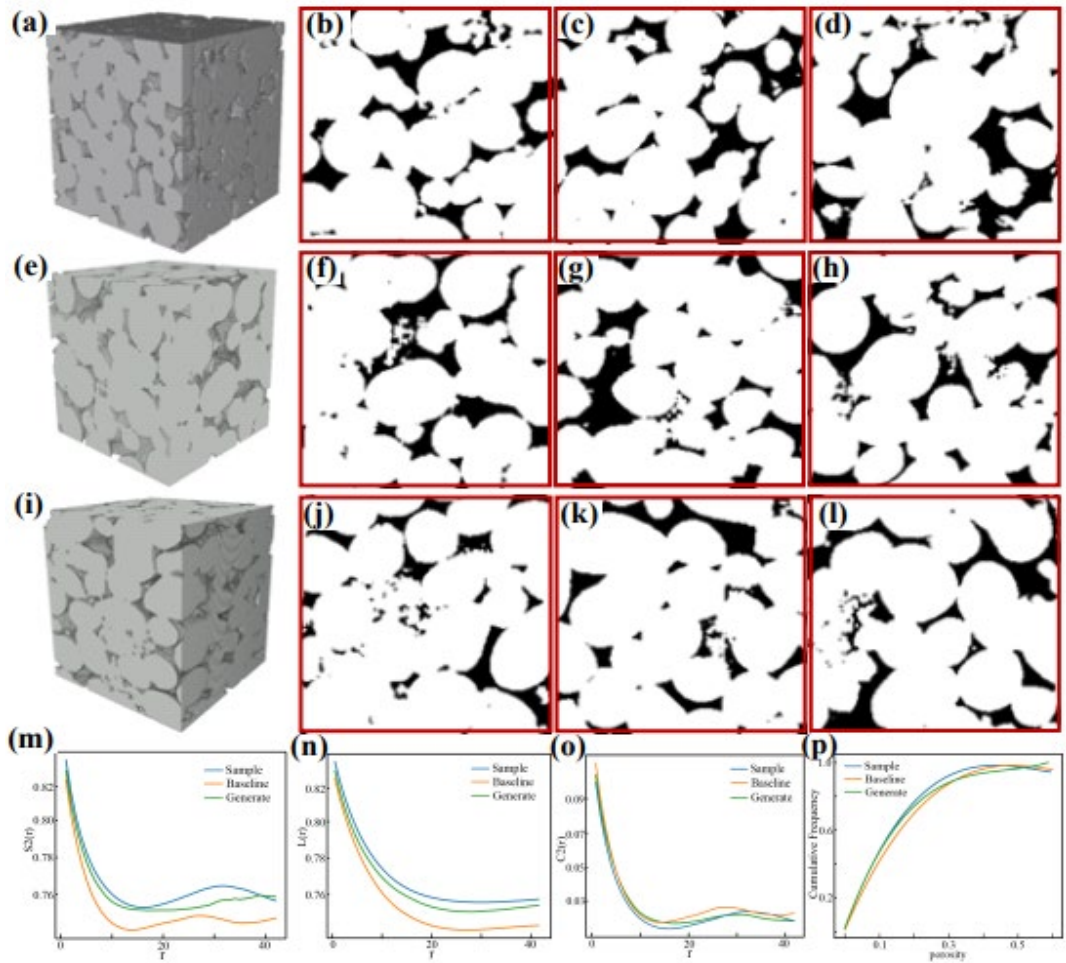

Fig. 4 2D and 3D visualizations of representative Ketton rock samples and experimental statistical results. Where a-d, e-h and i-l are the 3D visualizations of Sample, Baseline and our model, as well as the slices along the different directions of X, Y and Z respectively. Sections m-p are the results of comparison of the average along the different directions $S2(r)$, $C2(r)$, $L(r)$ and the cumulative distribution of local porosity along different directions (real samples in blue, Baseline samples in yellow, and generated samples in green)

## 2.3 Case 3: Lithium battery NMC cathode material

To validate the generalization capability of the model proposed in this study in isotropic tri-phasic materials, the NMC cathode material of lithium batteries was selected as the test subject. As illustrated in Figure 5, all three phases exhibit complex geometric morphologies and demonstrate significant randomness in their spatial distribution. Considering that the reconstructed 3D structures are visually indistinguishable from the original samples, further comparative analysis was conducted using morphological parameters and visual indicators. The specific morphology descriptions are shown in **Supplementary Information S2**.

As seen in section m of Fig. 5, the generated tri-phasic average distribution is observed to be extremely similar to the real sample. Furthermore, the model exhibits a more stable isotropic distribution in the linear path function plot n compared to the baseline method. Accordingly, we can confirm that the model successfully generates a 3D multiphase microstructure equivalent to the original sample. The corresponding 2D and 3D structures are visualized in Fig. 5, a–i. More importantly, for the tri-phasic microstructure of NMC, the relative diffusion coefficients, phase volume fractions and other related indexes are detailed. Their specific distribution in the different phases is shown in **Supplementary Information S5**. All of them show that the algorithmic model proposed in this study exhibits excellent accuracy in microscopic material mass generation.

In particular, experiments focusing on isotropic tri-phase Lithium battery NMC cathode (LiNiMnCoO2) samples are conducted in this paper. Detailed ablation experiments on the MHFA module, the MSCA module and the CBAM module are carried out. These experiments aim to understand the influencing roles of each module in the algorithm model, further validating the proposed model's generalization. For related experimental results, reference is made to Table 4, where the optimal results are shown in bold.

The results in the table show that the values of SSIM and PSNR improved to different degrees after the introduction of different modules. This highlights the importance of these modules in enhancing the network model's generation capability. Specifically, compared to the baseline model, when the model designed in this study further incorporated the MHFA, MSCA and CBAM module the values of SSIM and PSNR reached 0.461 and 22.035. Due to the uniqueness of the material's microstructure, it is difficult to distinguish the generated effect of a single slice from the real sample morphologically. Therefore, the generation quality evaluation criteria are further combined for comparison. After a comprehensive evaluation, the generation quality and reconstruction performance of the model were shown to be the best when all modules were introduced. **Supplementary Information S6** shows in detail the comparison of the tri-phasic averaged SSIM, PSNR and baseline models for different planes of the lithium battery cathode NMC.

It should be noted that macroscopic image generation tasks often involve distinguishing between foreground and background information, so there may be certain areas of focus in the model. In contrast, material microstructures are unique in that they do not distinguish between foreground and background information. Therefore, the central goal of this task is to reproduce images that are as similar as possible to the original structure. Since the reconstructed structure has randomness, it is difficult to generate a similar structure in the real sample and the generated sample in the same layer of slices. The specific scheme of this algorithm calculates SSIM and PSNR for all slices of the image in the X, Y and Z planes at every layer. It then seeks the average of values for each slice that corresponds to the real sample to make comparisons.

Based on this, although the SSIM and PSNR calculated in this article are relatively low, they can also measure the generative ability of the model itself. To evaluate the generation quality of different materials, it is necessary to refer to the relevant metrics in this field, such as those in Fig. 5 and **Supplementary Information S6**. The detailed analysis of the ablation experiment results is shown in Supplementary

Information S6.

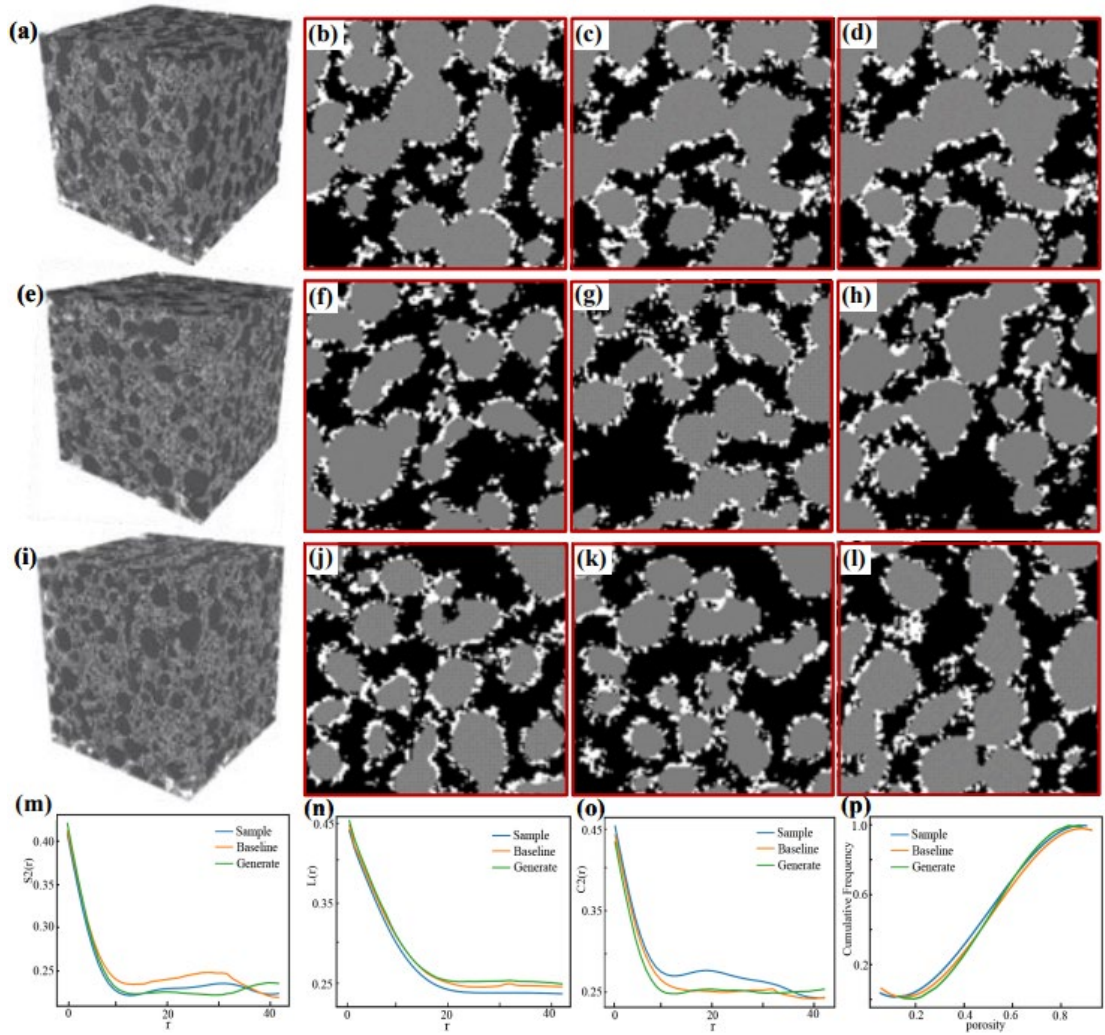

**Fig. 5 2D and 3D visualizations of representative isotropic tri-phase Lithium battery NMC cathode samples and experimental statistical results. Where a-d, e-h and i-l are the 3D visualizations of Sample, Baseline and our model, as well as the slices along the different directions of X, Y and Z respectively. Sections m-p are the results of comparison of the average along the different directions $S2(r)$, $C2(r)$, $L(r)$ and the cumulative distribution of local porosity along different directions (real samples in blue, Baseline samples in yellow, and generated samples in green)**

**Table 4 Comparison of ablation experiments for MHFA module, MSCA module and CBAM module**

| Baseline | MHFA | MSCA | CBAM | SSIM | PSNR |
|---|---|---|---|---|---|
| √ |   |   |   | 0.266 | 18.493 |
| √ | √ |   |   | 0.312 | 20.357 |
| √ |   | √ |   | 0.290 | 18.984 |
| √ |   |   | √ | 0.286 | 18.571 |
| √ | √ | √ |   | 0.301 | 18.936 |
| √ | √ |   | √ | 0.369 | 20.956 |
| √ |   | √ | √ | 0.347 | 19.203 |

## 2.4 Case 4：Hypoeutectic white cast iron

Fig. 6 i demonstrates the synthesized 3D structure, which has a high degree of visual consistency with the real sample a. Compared with the structure e generated by the baseline GAN network, the model in this paper is closer to the real sample in the 3D structure reconstruction. Although some observational effects may not be easily noticeable, an in-depth analysis by referring to the morphological descriptor metrics and the internal evaluation criteria of the model can lead to an accurate judgment. Furthermore, Fig. 6 reveals that the cross sections of the synthesized 3D structure observed from different directions of X, Y and Z all show a high degree of consistency with the test target in morphology. This not only verifies the authenticity and versatility of the model but also highlights the model's excellent ability in reconstructing topologically complex porous media.

Meanwhile, the m-p of Fig. 6 demonstrates the mean values of various morphological descriptors when reconstructing hypoeutectic white cast iron using the model of this paper. It can be observed that the reconstructed results (the green line) are close to the real samples in the $S2(r)$ and perfectly match the real data in terms of the $L(r)$.

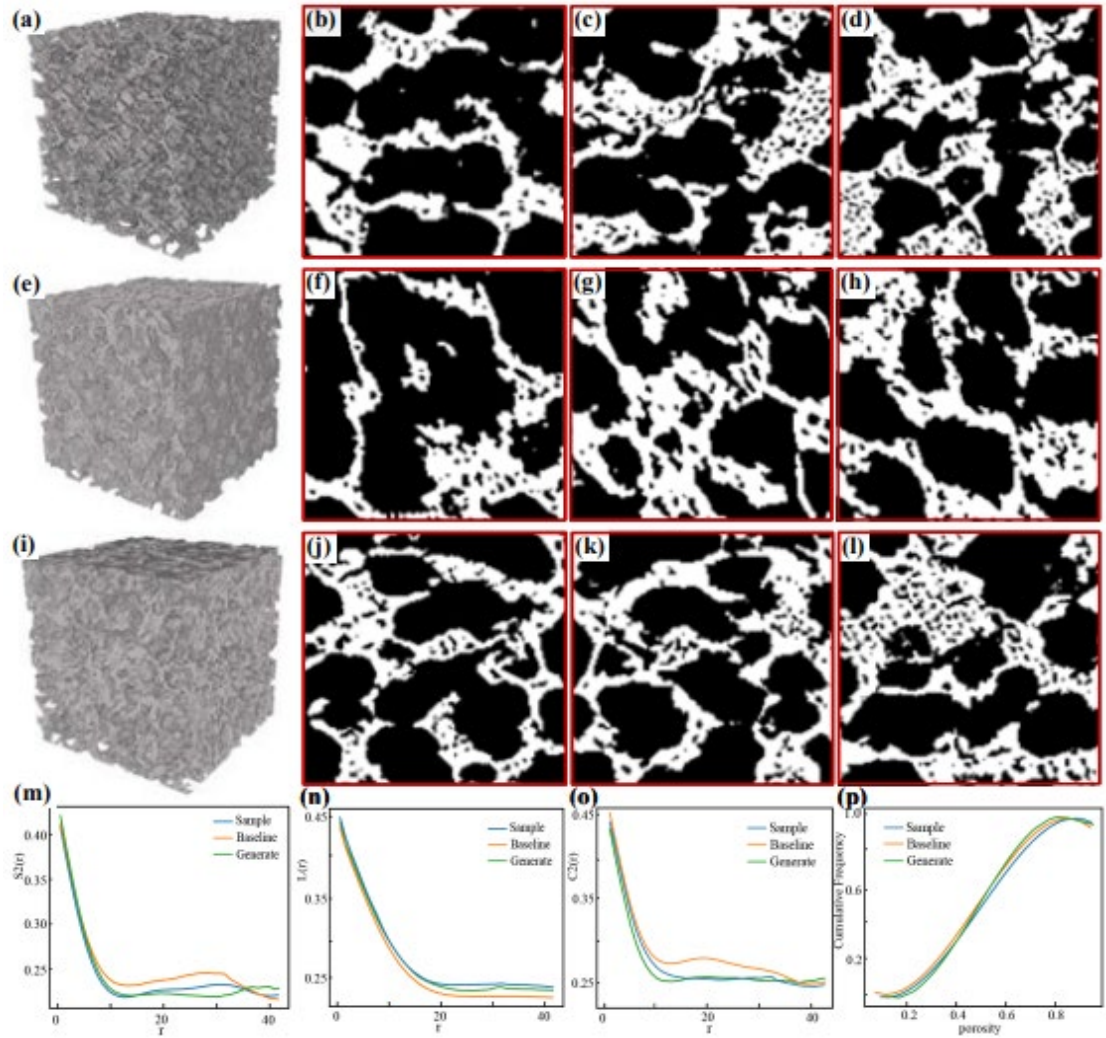

Fig. 6 2D and 3D visualizations of representative Hypoeutectic white cast iron samples and experimental statistical results. Where a-d, e-h and i-l are the 3D visualizations of Sample, Baseline and our model, as well as the slices along the different directions of X, Y and Z respectively. Sections m-p are the results of comparison of the average along the different directions $S2(r)$, $C2(r)$, $L(r)$ and the cumulative distribution of local porosity along different directions (real samples in blue, Baseline samples in yellow, and generated samples in green)

To accurately evaluate the performance of the proposed algorithm, the hypoeutectic white cast iron samples are also evaluated quantitatively. According to the data in Table 5, the average SSIM of the baseline between all the slices and real samples in the X, Y and Z planes is only 0.231, whereas our method improves to 0.351. In the PSNR evaluation, the benchmark network reaches 24.392, whereas our model is 26.336. Although these two metrics do not fully reveal the full quality of the generated material, they do demonstrate the performance advantages of the proposed algorithm. Combined with the aforementioned evaluation results, we can confidently assert that the algorithm proposed in this study is effective.

Table 5 Quantitative comparative metrics of SSIM, PSNR in Hypoeutectic white cast iron samples

| Method | Metrics | |
|---|---|---|
| | SSIM | PSNR |
| Baseline | 0.231 | 24.392 |
| OUR | **0.351** | **26.336** |

### 2.5 Case 5：Copper-zinc alloy

Lastly, the study choose Copper-zinc alloy samples with obvious anisotropy in the image for generalization experiments. Figure 7 shows the reconstructed 3D structures. Compared to the Baseline (the GAN network) across various directional average generation quality metrics, they bear a striking resemblance to the original samples and demonstrate considerable diversity. This result confirms the advantage of our model in reconstructing the anisotropic materials. Meanwhile, considering the difference in the properties of anisotropic materials in different directions, we show the average results of the $S2(r)$, $C2(r)$, $L(r)$ and the cumulative distribution of localized pores in the X, Y and Z directions in detail in **Supplementary Information S7**. It is remarkable to see that the results along the Z-axis are slightly lower than the other orientations of the structure. However, compared to the baseline generated results on the Z-axis, it performs well in terms of generation quality and continuity. This is attributed to the 3D convolutional kernel used by the model for feature extraction, which allows features in the Z-axis direction to be processed efficiently. In addition, Fig. 7 sections i-l illustrate the resultant plots of the generated 2D slices and 3D structures. When compared with the real samples shown in sections a-d, these plots reveal that the model's 2D slices and 3D structures are highly consistent with the measured samples. This consistency is observed both in terms of the cumulative distribution of localized pores and the morphology of the overall 3D structures. This outstanding performance can be attributed to the generator of the algorithm model, which is able to accurately capture and reconstruct these complex feature structures.

To further quantitatively evaluate the effectiveness of the model, we calculated the average of the evaluation indicators in different directions to more comprehensively evaluate the overall generation performance of the model. As shown in Table 6, the proposed model exhibits better results in terms of both average SSIM and PSNR when compared to baseline.

Table 6 Quantitative comparison indexes of SSIM and PSNR in Copper-zinc alloy samples

| Method | Metrics | |
|---|---|---|
| | SSIM | PSNR |
| Baseline | 0.100 | 23.247 |
| OUR | **0.201** | **24.681** |

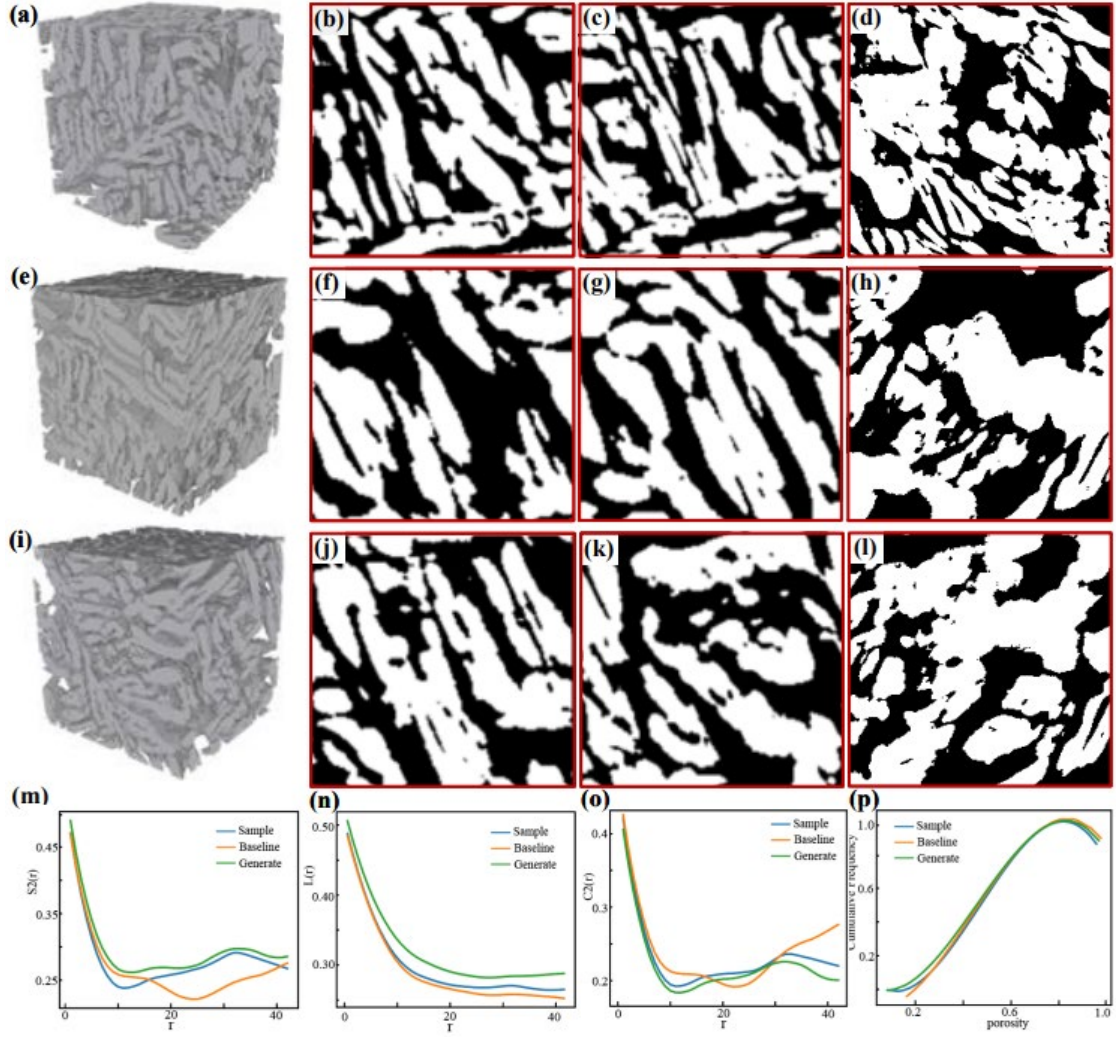

**Fig. 7 2D and 3D visualizations of representative anisotropy Copper-zinc alloy samples and experimental statistical results. Where a-d, e-h and i-l are the 3D visualizations of Sample, Baseline and our model, as well as the slices along the different directions of X, Y and Z respectively. Sections m-p are the results of comparison of the average along the different directions S2(r), C2(r), L(r) and the cumulative distribution of local porosity along different directions (real samples in blue, Baseline samples in yellow, and generated samples in green)**

Meanwhile, in this research, an ablation study was conducted on the loss function formulated for the network using a lithium battery cathode NMC. As shown in Table 7, the experimental results objectively confirm that the proposed loss function in this study enhances the performance constraints of the algorithm network. More specifically, when the image regularization loss $L_1(Y, G)$ is combined with the Wasserstein loss, the model exhibits superior generative capabilities compared to the Wasserstein loss alone. This combined loss strategy results in an SSIM value of 0.461, while the PSNR metric also achieves an improvement of 1.164.

**Table 7 Comparison of comparison results using different losses**

| Loss | Metrics | |
|---|---|---|
| | SSIM | PSNR |
| Wasserstein loss | 0.453 | 20.871 |
| Wasserstein loss+ $L_1(Y, G)$ | **0.461** | **22.035** |

Finally, in this study, macroscopic printing of the microstructure of synthetic materials has been successfully achieved using a Creality CR-10 Max model 3D printer. The key to this process lies in converting the fine microstructure to a macroscopic scale, facilitating a more in-depth analysis of the material's internal structure. As observed in the images in **Supplementary Information S8**, the 3D-printed synthetic materials are not only colorful in detail but also maintain a high degree of visual consistency with the original materials. This experimental result demonstrates the effectiveness of the employed algorithmic model in reconstructing the specified material samples. It highlights the model's potential for practical application in the manufacturing field, as detailed in Supplementary Information S8.

## 3 Conclusions

This study proposes an advanced generative model that fuses U-net and GAN. It is designed for efficiently capturing the complex features of microstructures and obtaining better quality 3D reconstruction and performance analysis from 2D images. The adversarial training strategy of GAN and the multi-scale property of U-net are fully integrated in this network, while the image regularization loss is combined with the Wasserstein distance loss. This integration enables the generation of isotropic and anisotropic porous media in a stable and efficient manner. Furthermore, this model automatically extracting the structural features of an image and reconstructing the 3D structure of a material using only one 3D training sample. Through experimental validation, the model has demonstrated superiority and generalization in the microstructure reconstruction of homogeneous/inhomogeneous materials.

## 4 Materials and Methods

### 4.1 Multi-scale hierarchical feature aggregation module

MHFA module is embedded between each 3D deconvolutional layer of the generative network, as shown in Fig. 1. The uniqueness of microstructure images requires that a high level of attention be given to subtle features, which determines that the model needs to extract different sensory field features. For this reason, the MHFA module initially employs 3×3, 5×5 and 7×7 convolutions to capture the contextual information of microstructural features at different scales. Subsequently, to enhance the robustness and stability of the model, the MHFA module is incorporated with a dense hierarchical feature connection strategy. The dense hierarchical feature not only is found to enhance the feature representation of semantic and geometric structures but also improves the richness and efficiency of feature extraction compared to traditional GAN networks. Specifically, this dense hierarchical feature connection comprises four 3×3 convolutional layers and one 1×1 convolutional layer. Features with different receptive fields introduced in the second and third layers effectively optimize the feature extraction and utilization efficiency, as illustrated in Fig. 2b.

### 4.2 Multi-scale channel aggregation module

Integrating the MHFA module after each convolutional layer extends the sensory field of the image features. This enhancement of the representation of the local microstructure details in the overall features brings them closer to the real structural features. The proposed MSCA module further enhance the representation of geometric and semantic features, which mainly incorporates multi-scale convolution strategies including 3×3, 5×5 and 7×7. This module aims to capture the contextual detail features of the microstructures, allowing for the overlay of different feature scales in the low-resolution feature layers. For instance, in reconstructing the battery cathode material structure, it is possible to reveal the distribution of features between different phases. After extracting these features, the model focuses on strengthening the inter-channel feature correlation to improve network expressiveness and model generalization. To this end, the channel attention mechanism (CAM) module is introduced after the multi-scale convolution. The CAM module can screen out the non-critical features. This reduction the model complexity and computational cost enhances the training efficiency of algorithm model. The structural design is demonstrated in Fig. 2a.

As illustrated in Figure 2c, the CAM module implements global maximum pooling and average pooling at the channel level. Maximum pooling extracts the most salient features within a single channel, while average pooling contributes to preserving feature invariance and smoothing. Subsequently, the results of the two pooling are spliced at the channel level. Following one convolution process, the number of channels is reduced to one. A spatial attention feature is then generated via a sigmoid function and eventually multiplied with the input features of the CAM module to produce output. As shown in Equation (1).

$$CA(F) = sigmoid(MLP(AvgPool(F)) + MLP(MaxPool(F)))$$
$$= sigmoid(W_0(W_1(F^c_{Avg})) + W_0(W_1(F^c_{Max}))) \quad (1)$$

Where $F$ is the 3D feature, $Sigmoid$ is the activation function, $MLP$ is the multilayer perceptron, $F^c_{Avg}$ is the average pooled feature, $F^c_{Max}$ is the maximum pooled feature, and $W_0$、$W_1$ denote the neural network weights.

## 4.3 Convolutional block attention module

At the primary stage of the generative network, the 3D features of the image are successfully extracted. To more accurately model the microstructure of the original material and to enhance attention to the spatial and channel relationships in the deeper layers of the network, a composite attention module, called CBAM, is introduced in the end layers of the generator (the last four to five layers). As demonstrated in Figs. 2c and 2d, it consists of two independent sub-modules: the CAM and the Spatial Attention Mechanism (SAM). Weights are applied to the channel and spatial dimensions by these two submodules respectively, to maintain favorable feature information while screening out unnecessary features. Furthermore, to optimize the computational complexity of the model in time and space, the SAM module is cascaded after the CAM module. Specifically, MaxPool and AvgPool are executed for the channel dimension to generate two feature maps respectively. Subsequently, these two features are collocated along the channel dimension to perform a stitching operation, further a 7×7 convolution is applied, before the features are refined to a single channel through a dimensionality reduction technique. Finally, a sigmoid function is applied to analyze the correlation between spatial elements to form spatial weights. This operation enhances the robustness and stability of the model when dealing with complex inputs, and its mathematical expression is shown in Equation (2).

$$SA(F) = sigmoid(Conv3D^{7\times7\times7}([AvgPool(F); MaxPool(F)])) \quad (2)$$

where $F$ is a 3D feature and $Conv3D^{7\times7\times7}$ indicates that this feature is generated by a convolutional layer with a convolutional kernel of 7×7×7.

In microstructure images, the distinction between foreground and background is not emphasized, and all features in the image are considered of equal importance. Therefore, greater attention to the global information is required by the model, a unique approach is implemented in the discriminator to specifically evaluate the 'realism' of the image at the 2D level. Specifically, inspired by Steve et al. [58], the 3D microscopic images are randomly sliced with equal dimensions along the X, Y and Z cross-sectional directions, and the same slicing strategy is adopted during network training. This approach not only reduces the number of network parameters relative to the 3D discriminator, but more importantly, strengthens the focus on all feature information within microstructure images, which improves the judgment accuracy of the model. The related network structure in Table 1.

**Table 1 Differential structure**

| Layer | Structure | Parameters |
|---|---|---|
| Layer1 | Batch Normalization<br>ReLU<br>Conv2D | In_channel, out_channel = (32, 64)<br>Kernel_size=4<br>Stride=2 padding=1 |
| Layer2 | Batch Normalization<br>ReLU<br>Conv2D | In_channel, out_channel = (64, 128)<br>Kernel_size=4<br>Stride=2 padding=1 |
| Layer3 | Batch Normalization<br>ReLU<br>Conv2D | In_channel, out_channel = (128, 256)<br>Kernel_size=4<br>Stride=2 padding=1 |
| Layer4 | Batch Normalization<br>ReLU<br>Conv2D | In_channel, out_channel = (256, 512)<br>Kernel_size=4<br>Stride=2 padding=1 |
| Layer5 | Batch Normalization<br>ReLU<br>Conv2D | In_channel, out_channel = (512, 1)<br>Kernel_size=4<br>Stride=2 padding=0 |

### 4.4 Loss function

The loss function, a critical component of the network model, measures the difference between the model predictions and the actual labels. It is minimized by adjusting the model parameter settings to bring the model output closer to the actual observations. Especially in GAN models, the principle of the loss function can be explained from the viewpoint of Kullback-Leibler divergence and Jensen-Shannon divergence, as seen in Equation (3).

$$D_{KL}(P \| Q) = E_{x \sim p} \left[ \log \frac{P(x)}{Q(x)} \right] \tag{3}$$

Here, the probability of a random variable taking x in the probability distributions $P$ and $Q$ is denoted by $P(x)$ and $Q(x)$ respectively, while the expected value of a randomly sampled $x$ under the probability distribution $P$ is denoted by $E_{x \sim P}$ It is worth noting that $D_{KL}(P\|Q)$ due to its asymmetry, is inconsistent in measuring the difference between two probability distributions that may lead to the loss of information, which has an impact on the training and performance of the model. Therefore, this research employs $JS$ divergence to quantify the average similarity between two distributions, which is defined as follows:

$$D_{JS}(P \| Q) = \frac{1}{2} D_{KL}(P \| M) + \frac{1}{2} D_{KL}(Q \| M) \tag{4}$$

where the mean distribution of the distributions $P$ and $Q$ is denoted by $M$, with $M$ being $0.5(P + Q)$. $D_{KL}(P\|M)$ denotes the $KL$ divergence between the distribution $P$ and the mean distribution, and $D_{KL}(Q\|M)$ denotes the $KL$ divergence between the distribution $Q$ and the mean distribution $M$. The value of the $JS$ divergence is always between 0 and 1.

Within the original GAN architecture, the discriminator is used to distinguish between the real data distribution $P_{data}$ samples and the generator samples $P_g$, while the generator works to make $P_g$ closer to $P_{data}$. When the model training reaches convergence, the discriminator strives to maximize the $JS$ divergence between the real data and the generated data, while the generator seeks to minimize the $JS$ divergence. Thus, the loss function of the original GAN is shown in Equation (5).

$$min_G max_D V(D,G) = E_{x \sim p_{data}}[\log D(x)] + E_{z \sim p_z}[\log(1 - D(G(z)))] \tag{5}$$

where $min_G$ and $max_D$ denote a two-stage optimization process. First, for a fixed generator $G$, it is desired to maximize the performance of the discriminator $D$. Subsequently, with the optimal discriminator determined, the objective shifts to minimizing the loss of the generator $G$. $V(D, G)$ is the GAN objective function. $E_{x \sim pdata}$ denotes the expected value of the sampled data points $x$ in the computed real data distribution $P_{data}$ and $E_{Z\sim p_Z}$ denotes the expected value of the sampled random noise $z$ in the prior distribution $P_Z$; $G(z)$ describes the data points generated by the generator based on the random noise $z$. $logD(x)$ is the logarithmic probability of the "true" label assigned by the discriminator $D$ to the true data point $x$. While $log(1-D(G(z)))$ denotes the logarithmic probability of the "false" label assigned by the discriminator $D$ to the data point generated by the generator $G$.

However, when the generator sample data $P_g$ appears to be non-overlapping or the overlap is negligible with the real data $P_{data}$, then the *JS* scatter will no longer be effective, which can cause the generator to suffer from the gradient vanishing problem. In addition, the original GAN model may face several challenges, including training difficulties, pattern collapse, insufficient diversity of generator samples, and loss functions that cannot effectively guide the network training process. Therefore, Our algorithm introduces the Wasserstein distance (also known as Earth Mover's Distance, EMD) to replace the *JS* scatter used in traditional GAN. This scheme not only measures the difference between the true and generated distributions but also enhances the stability and convergence of model training by ensuring that the discriminator satisfies the 1-Lipschitz continuity constraint. To achieve this goal, after each model update, the algorithm employs a weight trimming strategy that restricts the absolute value of the weights to no more than a predefined constant C. With these improvements, the algorithm model optimizes the loss function of the original GAN, as described in Eq. (6).

$$min_G max_D V(D,G) = E_{x\sim p_{data}}\left[D(x)\right] + E_{z\sim p_z}\left[D(G(z))\right] \quad (6)$$

where $E_{x\sim pdata}[D(x)]$ denotes the expected value of the data x sampled from the true distribution $P_{data}$ evaluated by the discriminator $D$. $E_{Z\sim P_Z}[D(G(z))]$ denotes the expected value of the random noise $z$ sampled from the a priori distribution $P_Z$, transformed by the generator $G$, and then evaluated by the discriminator $D$.

Meanwhile, the *L1*-paradigm loss (also known as absolute error loss) and *L2*-paradigm loss (i.e., mean-square error loss) are widely used in the field of image generation, and formulas are shown in (7) and (8).

$$L_1(y, f(x_i)) = \sum_i |y_i - f(x_i)| \quad (7)$$

$$L_2(y, f(x_i)) = \sum_i (y_i - f(x_i))^2 \quad (8)$$

Here, $y_i$ is the target value, and $f(x_i)$ is the estimated value. Inspired by Yu [59] and Isola [60] et al., $L_1$ paradigm loss performs better in improving robustness and reducing image blurring. Therefore, the $L_1$ paradigm loss is introduced to motivate the computation of the generator $G$. The aim is to reduce the pixel-level difference between the generated multidirectional sliced image and the original structural sliced image, thus to generate a 3D structure that is closer to the actual one. This loss is referred to as the image regularization loss, see Equation (9).

$$L_1(Y,G) = \sum_i |Y - G| \quad (9)$$

where $G$ is the generated slice image and $Y$ is the slice image of the original sample.

In summary, this algorithm designs the above two losses as the total loss function of this model as shown in Equation (10).

$$L_{total} = \lambda_w \cdot min_G max_D V(D,G) + \lambda_r \cdot L_1(Y,G) \quad (10)$$

where $\lambda_w$、$\lambda_r$ are Wasserstein distance loss weights and image regularization loss weights respectively.

**4.5 Experimental equipment**

In this study, the proposed algorithmic model was trained and tested on the Ubuntu 20.04 operating system by employing the Pytorch deep learning framework. The experimental environment configuration includes an Intel(R) Xeon(R) Platinum 8175M CPU @ 2.50GH and an NVIDIA GeForce RTX 3090 quad GPU with 24GB of memory. During the algorithm optimization process, the Adam optimizer was selected, a uniform learning rate was set for both the generative and discriminative networks. In addition, the model was iteratively updated, with $\beta_1$ and $\beta_2$ set to 0.9 and 0.999 respectively. For the specific implementation, the training batch size was set to 16 and 100 training cycles were performed until the model converged.

**4.6 Datasets**

To comprehensively evaluate the effectiveness and generality of the algorithmic model proposed in this study, image samples of five materials were carefully selected for comprehensive analysis. Representative data samples from different fields were selected using the anisotropy index: berea sandstone, ketton rock, lithium battery NMC cathode, hypoeutectic white cast iron and copper-zinc alloy. Meanwhile, in the process of data sample selection, it was observed that in the field of materials, isotropy and anisotropy refer to the variation of material properties with direction. In contrast, in the field of imaging, isotropy and anisotropy are used to characterize the features of images or the image processing algorithms.

For example, in the work of Steve et al. [58], a distinction was not made between the isotropy / anisotropy of the material and the isotropy / anisotropy of the image, which can lead to errors in performance analysis when analyzing the generative properties of the material. Based on this problem, we have made a detailed selection of the analyzed data set in **Supplementary Information S1**. In the selected data samples, it is ensured that the algorithm proposed in this study is able to reconstruct 3D structures that not only help to analyze the material properties in depth but also reduce the cost effectively. See **Supplementary Information S2** for the specific data presentation. Meanwhile, the image boundary setting is equally important to the reconstruction quality, and the method of image boundary setting in this paper is shown in **Supplementary Information S3**.

**4.7 Evaluation Metrics**

**4.7.1 Evaluation of the quality of generation**

In order to accurately assess the performance and quality of the model-generated samples, metrics such as the two-point correlation function $S_2(r)$, the two-point clustering function $C_2(r)$, the linear path function $L(r)$ (where $r$ denotes the distance between the two points) and the cumulative distribution of the local porosities were introduced. These metrics permit an in-depth analysis of the complex properties of the material's microstructure at the microscopic level. Detailed definitions and formulas are shown in **Supplementary Information S4**

In particular, three key qualitative evaluation metrics for NMC cathode materials for lithium batteries in the electrochemical field are introduced in this research: phase volume fraction, specific surface area (SSA), triple phase boundary (TPB) and relative diffusion coefficients. The definitions and mathematical formulas of these metrics are shown in **Supplementary Information S4.**

**4.7.2 Evaluation of network models**

To quantitatively assess the performance of the algorithmic models, the Structural Similarity Index (SSIM) and Peak Signal-to-Noise Ratio (PSNR) evaluation metrics have been introduced [61], enabling the analysis of visual similarity and dissimilarity between images at a macroscopic level. Specifically, SSIM, which aims to measure the visual similarity between two images, is calculated as shown in Equation (11).

$$SSIM(x,y) = \frac{(2\mu_x\mu_y + C_1)(2\sigma_{xy} + C_2)}{(\mu_x^2 + \mu_y^2 + c_1)(\sigma_x^2 + \sigma_y^2 + c_2)} \quad (11)$$

where $\mu_x$ and $\mu_y$ are the mean values of the generated image $I_1$ and the original $I_2$ within window $x$, $\sigma_x^2$ and $\sigma_y^2$ are the variances of the images $I_1$ and $I_2$ within window $x$, $C_1$ and $C_2$ are constants designed to avoid zero denominators and are typically $C_1=(K_1L)^2$ and $C_2=(K_2L)^2$ where L is the dynamic range of the pixel and $K_1\ll1$、$K_2\ll1$ are non-zero constants. This metric responds to the correlation between pixel points within a certain sliding window, with the calculation of brightness, contrast and structural attributes between the images in a localized range. These attribute metrics are then averaged. Therefore, in this algorithm, the sum of all the mean values of SSIM of the generated slices of the structure along the different planes of X, Y and Z is compared with the real corresponding sliced image, which can reflect the similarity between the generated image and the real image to a certain extent. When the SSIM value is closer to 1, it means that the images are more similar, thus proving that the network model is more advanced. At the same time, this index is able to reflect only the model itself and not the quality of the generated material properties. Therefore, it is necessary to combine performance indicators of different materials to observe the quality of material generation. Sufficient experiments and explanations are provided in Case 4 battery cathode material reconstruction and **Supplementary Information S5**.

Mean Square Error (MSE) measures the mean square pixel difference between two images, which is defined as (12) given two images $I_1$ and $I_2$ of the same size:

$$MSE(I_1, I_2) = \frac{1}{MN}\sum_{i=1}^{M}\sum_{j=1}^{N}(I_1(i,j) - I_2(i,j))^2 \quad (12)$$

Where $(i, j)$ denotes the positional coordinates of the pixel point, $M$ and $N$ denote the height and width of the image. MSE is directly calculated as the average of the sum of squares of the difference of each pixel between the two images. According to MSE, it is known that PSNR is calculated based on the mean square error MSE, given an original image $I_2$ and a reconstructed image $I_1$ see Equation (13).

$$PSNR = 10\times\log_{10}\frac{MAX_I^2}{MSE} \quad (13)$$

Where $MAX_I$ is the maximum value of pixels in the image, MSE is the mean square error, a higher PSNR value indicates a smaller difference between the reconstructed image and the original image, thus reflecting the higher quality of the model. Although the generated structure has a high degree of randomness, this paper seeks the overall average of its three directions for comparison and then combined with the quality indicators for comparison. The overall effect of the generation can be highly close to the real samples, so the SSIM and PSNR image metrics used in this algorithm are effective for evaluating the strengths and weaknesses of the network itself.

# Supplementary Information

# A Generative Machine Learning Model for Material Microstructure 3D Reconstruction and Performance Evaluation

Yilin Zheng[1], Zhigong Song[1*]

School of Mechanical Engineering, University of Jiangnan 214122 Wu Xi

song_jnu@jiangnan.edu.cn

**S1: Datasets selection**

To comprehensively verify the effectiveness and wide adaptability of the algorithmic models in this study, five types of material data from different fields were selected for experimental testing. In the process of data selection, special consideration is given to the pragmatic requirements of engineering applications, ensuring that the algorithm is capable of reconstructing 3D structures. These structures are not only instrumental in facilitating a comprehensive analysis of material properties but also play a pivotal role in reducing costs in an efficacious manner. Furthermore, to demonstrate the versatility and generalization of the model based on the characteristics of data diversity, the homogeneity analysis results in the field of image analysis were utilized, and five classic image data samples were carefully selected. Morphologically, the homogeneity and dissimilarity of each image can be intuitively distinguished. The quantification of isotropy and anisotropy in homogeneity analysis is achieved by calculating the Gray-Level Co-occurrence Matrix (GLCM) [1], a statistical method that measures the variation of co-occurrence between pixel gray levels in an image with direction. Features of the GLCM such as contrast, homogeneity, energy, and entropy [2] are utilized to describe the isotropy or anisotropy of an image.

In GLCM, contrast reflects the distribution of pixel-to-gray scale differences in an image, serving as a measure of local variation. For anisotropic images, higher contrast is observed in certain directions, as shown in Equation (S1).

$$Contrast = \sum_{i,j=0}^{levels-1} P(i,j)(i-j)^2 \tag{S1}$$

Where $P(i, j)$ is the probability that the gray values between two pixels are i and j and have a particular spatial relationship, levels are the number of gray levels in the image.

Homogeneity also known as consistency or local homogeneity, reflects the smoothness and consistency of the image texture. For isotropic images, the homogeneity feature will be similar in different directions, see Equation (S2) for details.

$$Homogeneity = \sum_{i,j=0}^{levels-1} \frac{P(i,j)}{1+(i-j)^2} \tag{S2}$$

Energy known as the angular second moment, is a measure of the sum of squares of the element values in the GLCM, which reflects the uniformity and repeatability of the image texture.

$$Energy = \sum_{i,j=0}^{levels-1} P(i,j)^2 \tag{S3}$$

Entropy is a measure of the amount of information in an image, reflecting the complexity and disorder of the image texture. It will reach its maximum when all $P(i, j)$ are equal. The higher the entropy, the more irregular the texture distribution in the image and the stronger the anisotropy.

$$Entropy = -\sum_{i,j=0}^{levels-1} P(i,j)\log(P(i,j)) \tag{S4}$$

When $P(i, j)$ is 0, $0 \times \log(0) = 0$ is defined. These feature calculations will consider pixel pairs in specific directions (such as 0°, 45°, 90° and 135°) and distances. To reflect the characteristics of image texture in different directions. By comparing the values of these features in different directions, we can quantify the anisotropy of the image. For example, an image with approximately equal contrast, homogeneity, energy and entropy values in all directions is said to be isotropic, while an image in which these values differ significantly in some directions exhibits anisotropy.

Among the five data samples selected in this study, texture features such as contrast, homogeneity, energy and entropy are computed for the slices of the 3D data samples along the three orthogonal directions (the X, Y and Z axes). These three directions are considered to adequately represent texture properties in 3D structures. The steps are as follows:

(1) Extracting features: For each direction of the slice, GLCM texture features (contrast, homogeneity, energy and entropy) are computed. For each direction, the mean or other statistics (e.g. median) of these features are obtained separately to obtain the texture characteristics that represent each direction. (2) Comparing features in different directions: For each texture feature, the statistics of X, Y and Z directions are compared. If these features are similar in all three directions, then the 3D microstructure may be isotropic. If the statistics of these features are significantly different in one or some directions, it indicates that the 3D microstructure exhibits anisotropy. (3) Analyzing results: If all texture features are consistent in all directions, it indicates that the 3D microstructure is isotropic. If one or more texture features are significantly different in different directions, it indicates that the structure is anisotropic, as shown in Tables S1–S5.

**Table S1 Contrast, homogeneity, energy and entropy values and image properties for different directions of Berea sandstone**

| Dataset | Directions | Contrast | Homogeneity | Energy | Entropy | Properties |
|---|---|---|---|---|---|---|
| Berea sandstone | X | 767.695 | 0.869 | 0.667 | 2.956 | Isotropy |
| | Y | 786.904 | 0.866 | 0.690 | 2.943 | |
| | Z | 801.069 | 0.863 | 0.701 | 2.950 | |

**Table S2 Contrast, homogeneity, energy and entropy values and image properties for different directions of Ketton Rock**

| Dataset | Directions | Contrast | Homogeneity | Energy | Entropy | Properties |
|---|---|---|---|---|---|---|

| | X | 1168.381 | 0.856 | 0.723 | 2.952 | |
|---|---|---|---|---|---|---|
| Ketton Rock | Y | 1097.543 | 0.863 | 0.729 | 2.840 | Isotropy |
| | Z | 1153.762 | 0.857 | 0.723 | 2.945 | |

**Table S3 Contrast, homogeneity, energy and entropy values and image properties for different directions of Lithium battery NMC cathode**

| Dataset | Directions | Contrast | Homogeneity | Energy | Entropy | Properties |
|---|---|---|---|---|---|---|
| | X | 1589.011 | 0.681 | 0.449 | 6.175 | |
| Lithium battery NMC cathode | Y | 1531.274 | 0.693 | 0.460 | 5.983 | Isotropy |
| | Z | 1480.242 | 0.696 | 0.461 | 5.941 | |

**Table S4 Contrast, homogeneity, energy and entropy values and image properties for different directions of Hypoeutectic white cast iron**

| Dataset | Directions | Contrast | Homogeneity | Energy | Entropy | Properties |
|---|---|---|---|---|---|---|
| | X | 1639.290 | 0.718 | 0.507 | 5.239 | |
| Hypoeutectic white cast iron | Y | 1461.695 | 0.756 | 0.597 | 4.578 | Isotropy |
| | Z | 1587.715 | 0.732 | 0.552 | 4.978 | |

**Table S5 Contrast, homogeneity, energy and entropy values and image properties for different directions of Copper-zinc alloy**

| Dataset | Directions | Contrast | Homogeneity | Energy | Entropy | Properties |
|---|---|---|---|---|---|---|
| | X | 0.070 | 0.965 | 0.665 | 1.352 | |
| Copper-zinc alloy | Y | 0.083 | 0.959 | 0.677 | 1.353 | Anisotropy |
| | Z | 2697.677 | 0.749 | 0.496 | 4.724 | |

Based on the contrast, homogeneity, energy and entropy in Tables S1–S5, we utilize the anisotropy index (AI) [3-4] to quantify the extent to which the image properties of a material vary in different directions, as shown in Eq. (S5).

$$AI = \sqrt{\sigma_C^2 + \sigma_H^2 + \sigma_E^2 + \sigma_{En}^2} \tag{S5}$$

Where $\sigma_C$ is the standard deviation of contrast, which quantifies the degree of variability of light and dark differences between pixels in an image; $\sigma_H$ is the standard deviation of homogeneity, which describes the variation of consistency and homogeneity of the image texture; $\sigma_E$ is the standard deviation of energy, which expresses the repetitive or regular variability of the image texture; and $\sigma_{En}$ is the standard deviation of entropy, which measures the degree of dispersion of the image information in terms of uncertainty or complexity. The formula for the standard deviation is shown in equation (S6).

$$\sigma = \sqrt{\frac{1}{N-1} \sum_{i=1}^{N} (X_i - \overline{X})^2} \tag{S6}$$

where $\overline{X}$ is the mean value, to summarize, the AI formulation quantifies anisotropy by aggregating the directional variance of these descriptors:

When the Anisotropy Index (AI) approaches zero, it is indicated that the features exhibit a high degree of similarity across various directions, suggesting isotropy of the image. With a medium AI value, variations in the features across different directions are observed, yet these variations are not particularly significant, indicating a lack of significant anisotropy and thus suggesting isotropy. When the AI value is high, significant differences in features across various directions are indicated, denoting notable anisotropy in the image.

Table S6 clearly shows that the five data samples we have selected are strongly convincing, the anisotropy index of Berea sandstone is 16.738, that of Ketton Rock is 37.403, the lithium battery NMC cathode is 51.413, the hypoeutectic white cast iron is 91.354 and the copper-zinc alloy has an anisotropy index of 1556.5. Figure S1 demonstrates the results of the logarithmic values of the anisotropy indices for the five data sets. Specific morphological descriptions of the five data sets are presented in Table S7. Based on the anisotropy index, it is observed that the selected data sets are highly rational and representative. In cases where the logarithmic value of the anisotropy index exceeds 2.0 (indicating an anisotropy index of 100), the texture characteristics (contrast, homogeneity, energy, and entropy) of the sectioned images of the material's three-dimensional structure in each direction theoretically exhibit significant differences, classifying the characteristics of the images as anisotropic.

Table S6 Anisotropy index values for different datasets

| Datasets | Anisotropy Index | Properties |
| --- | --- | --- |
| Berea sandstone | 16.738 | Isotropy |
| Ketton Rock | 37.403 | Isotropy |
| Lithium battery NMC cathode | 51.413 | Isotropy |
| Hypoeutectic white cast iron | 91.354 | Isotropy |
| Copper-zinc alloy | 1556.5 | Anisotropy |

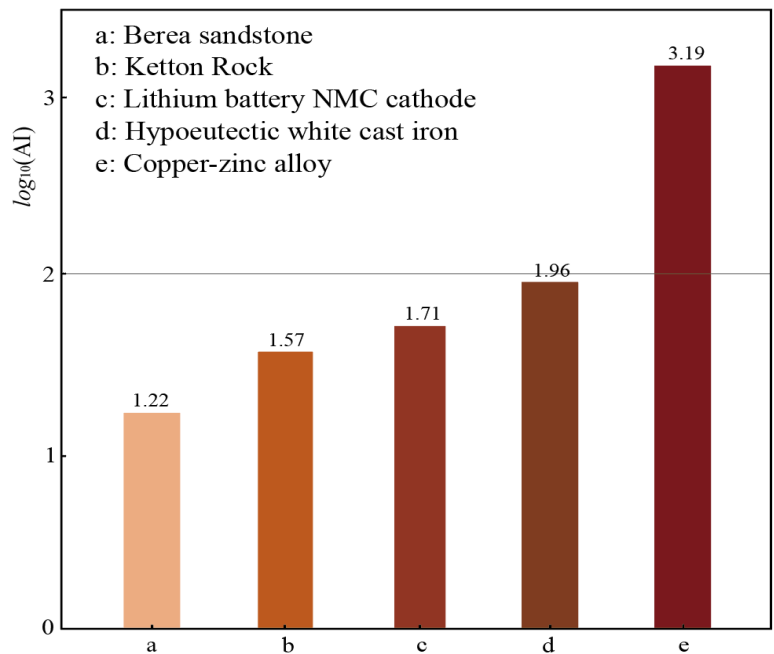

a: Berea sandstone
b: Ketton Rock
c: Lithium battery NMC cathode
d: Hypoeutectic white cast iron
e: Copper-zinc alloy

a: 1.22  b: 1.57  c: 1.71  d: 1.96  e: 3.19

**Fig. S1 Presentation of the results of taking logarithmic values of the anisotropy index for the five datasets**

**Table S7 Morphological descriptions specific to the five datasets**

| Descriptions | Datasets | Properties |
|---|---|---|
| Pronounced isotropy, high homogeneity and low contrast, relatively uniform distribution, more regular pore distribution | Berea sandstone | Isotropy |
| Anisotropy decreases but still maintains some degree of structural order, and the diversity of pore sizes and shapes begins to become more pronounced | Ketton Rock | Isotropy |
| General anisotropy, inconsistencies in particle size and shape, and more complex pore structures | Lithium battery NMC cathode | Isotropy |
| Show a relatively high degree of irregularity, but may still exhibit some anisotropy at the macro level | Hypoeutectic white cast iron | Isotropy |
| Pronounced anisotropy, with some degree of irregularity and variability in the microstructure | Copper-zinc alloy | Anisotropy |

## S2: Datasets introduction

### 1 Berea sandstone

Berea sandstone, shown in Fig. S2(a), is a sandstone from a river [5], whose structure consists mainly of individual sand grains cemented by clay minerals with cohesive properties and is therefore often chosen for sample reconstruction studies in porous media. Samples from quarries in the Berea area of Ohio, USA, were used in this study. The main components include quartz, feldspar, clay minerals, colluvium, and pores. The Berea sandstone is a fine to medium grained sandstone [6] that is relatively homogeneous in texture. Therefore, berea sandstone is frequently used in petroleum geology and engineering as a benchmark rock to explore the permeability and other physical properties of the rock. Additionally, it is commonly utilized in geological studies to simulate hydrocarbon reservoirs, thus providing insights into the storage and transport properties of hydrocarbons. For capturing the rock's macroscopic structure and the microscopic interactions between particles in detail, a reconstruction size of voxel size $252 \times 252 \times 252$ was chosen. It is worth noting that the reconstructed cross-section of Berea sandstone in the three orthogonal directions visually exhibits consistency and satisfies the property of isotropy.

### 2 Ketton Rock

The Ketton Rock sample, an oolitic limestone [7], is predominantly composed of 99.1% calcite and 0.9% quartz, exhibiting a granular composition ranging from spherical to ellipsoidal. In this sample, inter- and intra-granular pores, along with a large number of sub-resolution micropores that remain unresolved, are clearly observed. The presence of these sub-resolution pores enhances interactions with x-rays during the imaging process, resulting in a wide range of gray values within the grains, as shown in Fig. S2(b).

### 3 Lithium battery NMC cathode material

In the field of electrochemistry, the NMC (LiNiMnCoO2) cathode material [8] of lithium batteries, a key component, reveals its complex microstructure through open tomography data. This microstructure primarily consists of active materials, a conductive polymer binder containing carbon black and pores.

The central position occupied by the active material in the cathode is determinant in the battery's electrochemical performance, conferring excellent electrochemical properties, high energy density and a more stable cycle life. Carbon black, acting as the main conductive additive, synergizes with the polymer binder to ensure a strong bond between the active material particles and the current collector (e.g., aluminum foil), thereby enhancing the electrode's conductivity and ensuring stability during the charge/discharge cycle. Additionally, the pore structure provides essential paths for lithium ion transport from the electrolyte, playing a key role in the migration of lithium ions in the battery during charging and discharging.

The complete flow of the preparation method [9] covers the mixing of active materials, carbon black and binder with suitable solvents, which in turn forms a homogeneous slurry or printing ink. This slurry is applied on top of an aluminum foil (i.e., a current collector) using a coating technique, which is then placed in an oven for drying, and after drying the electrodes are further pressed or rolled to ensure their density and structural integrity. In this study, we utilized tri-phasic anisotropic Li-ion NMC cathode materials for experimental analyses and the relevant structural details are shown in Fig. S2(c).

4 Hypoeutectic white cast iron

Hypoeutectic white cast iron microstructure is mainly composed of ferrite and a smaller proportion of eutectic iron carbide ($Fe_3C$, also known as horizontal stone). When the carbon equivalent is lower than the critical value required for the formation of a stable eutectic phase, the combined amount of carbon and other alloying elements in the material is insufficient to maintain a stable eutectic state. As a result, iron carbide is produced in the form of flakes or needles dispersed within the ferritic matrix. This study utilizes the data of sub-eutectic white cast iron described in the literature [10] as an experimental sample see Fig. S2(d), the hard iron carbide in this cast iron contributes to the significant wear resistance. Compared with ordinary white cast iron, sub-eutectic white cast iron demonstrates better toughness and impact resistance. Through the analysis of this dataset, we are able to explore the intrinsic connection between the microstructure of the material and its macroscopic physical properties in a more in-depth manner, which is of great significance in directing the production of the relevant cast iron materials and their practical applications [11].

5 Copper-zinc alloy

The copper-zinc alloy Cu 60/Zn 40 (*wt*%) represents a brass of a specific composition having 60% copper and 40% zinc and is one of the typical compositions found in brass alloys [12-13]. The samples used in this study were subjected to an air cooling process, and the rapid cooling rate resulted in the formation of interwoven fibrous patterns on their surfaces, a phenomenon commonly referred to as the Weidmann structure. In this structure, the α-phase of the face-centered cubic structure is precipitated from the body-centered cubic *β*-phase (Cu-Zn solid solution) during the cooling process to form lamellar structures within a *β*-phase background. This lamination mainly aims to decrease the strain generated by the energy. It is noteworthy that the specific morphology and structure of the α and β phases are influenced by the cooling rate and heat treatment conditions. The reconstruction of their 3D microstructures using these data plays a crucial role in gaining an in-depth understanding of the mechanical properties and corrosion behavior of the materials [14-15], as detailed in Fig. S2(e).

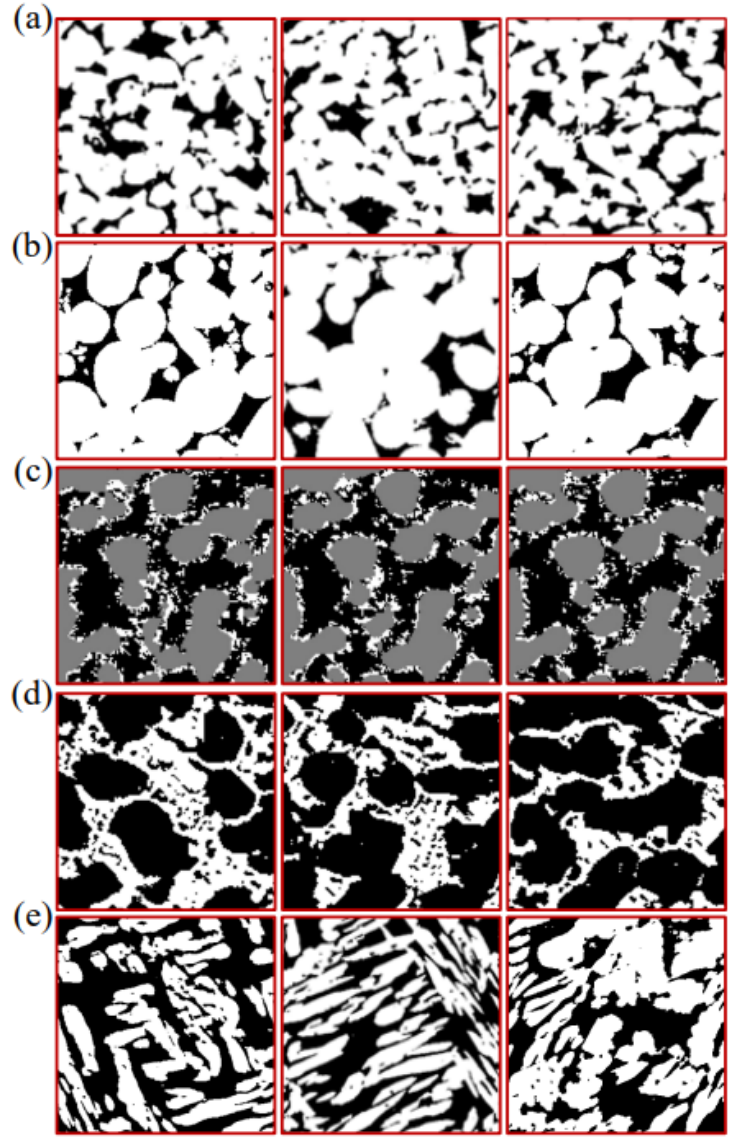

Fig. S2 Slices of different datasets along X, Y and Z directions (a is Berea sandstone, b is Ketton Rock, c is Lithium battery NMC cathode, d is Hypoeutectic white cast iron and e is Copper-zinc alloy)

### S3: Boundary setting

The setting of the boundaries significantly impacts the reconstruction, the voxels at the boundaries, which are the 3D space equivalents of pixels in an image, are not updated during the reconstruction process due to insufficient neighborhood information, remaining fixed throughout the process. While these boundary voxels provide neighborhood information, their lack of self-updating negatively affects the reconstruction quality. To minimize the boundary effect, periodic boundary conditions are utilized, wherein the values at one boundary of the image or voxel data are matched with the values at the opposite boundary, forming a continuous cycle. This approach ensures that the image acquires equivalent information at the boundary features and the features in the middle, which is applied in each layer of the reconstruction process to eliminate the boundary effect in subsequent steps.

Initialize the reconstruction layer: First, start with a 2D layer with random pattern boundaries, which is the starting point of the reconstruction process and whose boundaries are unoptimized and may contain distortions or discontinuities.

Removal of distorted boundaries: After the first round of reconstruction, the distorted boundaries of this layer are removed, meaning that the edge parts are eliminated and only the middle area is kept, which is considered to be more accurate and closer to the target structure.

Applying periodic boundaries: A periodic boundary is then added in the middle of this removed boundary, which means that one side of the boundary of the layer is "connected" to the opposite boundary. If the right boundary of the layer is observed, it will match the left boundary, creating a seamless cycle, which is intended to create a continuous structure and reduce edge effects.

Repeat the reconstruction process: Consider this newly processed 2D layer as an initial guess and perform the reconstruction procedure again. This step may be repeated to ensure that the boundary conditions are adequately applied and the reconstruction quality is optimized.

Reduction of boundary effects: By repeating this method once or twice, boundary effects can be significantly reduced, thus improving the overall quality of the reconstruction.

## S4: Evaluation formula of generation quality and Lithium battery NMC cathode material generation indexes and formulas

In materials science, the $S_2(r)$ is a commonly used tool that is defined as the probability of two points in space being simultaneously present in a particular phase, pore, or having a particular property. This function reveals the correlation of properties between two points in space and provides insight into microstructural features, see Equation (S7).

$$S_2(r) = \frac{1}{|V|} \int_V (I(x)I(x+r))dx \tag{S7}$$

where $r$ is the distance between two points in space, and $I(x)$ is an indicator function which is specific to a particular pore or phase in a given sample of rock or alloy, and the indicator function outputs a value of 1 if the selected location is in a pore or a particular phase; otherwise, it outputs a value of 0. $x$ and $x+r$ are two points in space respectively, $\int_V$ integrates over the entire domain $V$, when $r=0$, $S_2(0)$ is expressed as the medium's porosity.

The two-point cluster function $C_2(r)$ is a key metric used to assess reconstruction accuracy, which quantifies the spatial correlation between objects distributed in a random medium (e.g., particles or phases). Specifically, for any point in the medium, C2(r) defines the conditional probability of finding another point belonging to the same cluster or group at a distance r from that point, and the mathematical formulation of the function is shown in Equation (S8).

$$C_2(r) = \frac{\frac{1}{|V|}\int_V (I(x)I(x+r))dx}{(\frac{1}{|V|}\int_V (I(x)dx)^2} \tag{S8}$$

where $I(x)$ is the indicator function.

The linear path function $L(r)$ describes the average distance that a path can traverse until it encounters a different phase or feature (e.g., from nonporous to porous or from solid to air) as it proceeds from a random starting point in a random medium in a specified direction. In short, it evaluates the probability that a line segment of length r lies entirely within a pore phase, as shown in Eq. cf. (S9).

$$L(r) = Prob((I(x)=1, (I(x+1)=1, \cdots (I(x+r)=1)) \tag{S9}$$

where $I(x)$ is the indicator function, $I(x)=1$ is the origin, and $I(x+r)=1$ indicates that all points are in the pore phase. $L(r)$ allows for the assessment of the shape and size distribution of pores, grains, or other microscopic features. It provides a quantitative tool to assess the similarity of the reconstructed structure to the actual sample. In this study, we apply $L(r)$ to the Berea sandstone data samples. By comparing $L(r)$ of the original and reconstructed samples, we can not only assess whether the reconstructed samples accurately capture key pore structure features but also verify in the Cu-Zn alloy and hypoeutectic white cast iron data samples that the reconstructed structures are consistent with the distribution of grain sizes and morphologies in the actual samples. The closer the $L(r)$ of the reconstructed sample is to the actual sample, the better the quality of the reconstructed sample.

Local porosity is also a key indicator for assessing the quality of material structure generation, and for this purpose this study introduces a calculation method based on a 3D sliding window. First, the 3D volume data and the window size are taken as inputs. By moving the window size step by step over the entire 3D volume, all data points within the window are acquired, and their average values are calculated to obtain the porosity. To more accurately assess the generation quality of the algorithm, this metric is further post-processed in this paper. Specifically, the obtained local porosity values are sorted to plot their cumulative distribution function (CDF). The horizontal coordinate of the local porosity cumulative distribution function is the porosity value, and the vertical coordinate is the cumulative frequency of the observations less than or equal to the corresponding porosity value over the total observations. Obviously, the closer the generated curve is to the real data, the higher the quality of reconstruction.

Phase volume fraction reflects the proportion of a particular phase in the total material volume. A high volume fraction of a particular phase means that the properties of that phase (e.g., electrical conductivity or structural stability) will dominate the overall performance, as mathematically defined in (S10).

$$\phi_i = \frac{V_i}{V} \tag{S10}$$

where $V_i$ refers to the volume of phase $i$, calculated as a percentage of that corresponding voxel, $V$ denotes the total volume of the microstructural domain.

Specific surface area quantifies the surface area of a material per unit volume, closely relating to the electrochemical characteristics of the battery such as electric capacity and charge/discharge rate. Increased specific surface area provides a wider range of active surfaces, which potentially enhances the overall performance of the battery, which is given by (S11).

$$S_{V_i} = \frac{S_{A_i}}{V} = \frac{1}{V}\int dS_{A_i} \tag{S11}$$

where $S_{A_i}$ is the total surface area of the interface between phase $i$ and the remaining phases.

Triple phase boundary describes the intersection length between three phases $i$, $j$ and $k$ in a battery material, usually a region of three different encounter points, solid, gas, and electrolyte. In fuel cell applications, TPBs are critical for charge transfer and chemical reaction rates, and an increased number of TPBs is indicative of a greater number of active regions, which contributes to the improvement of the battery performance.

The relative diffusion coefficient (also known as the effective diffusion coefficient $D_{eff}$) is the actual diffusion coefficient of a substance in a porous medium, i.e, the ratio of the effective diffusion coefficient in a particular medium (e.g. porous medium) to the free diffusion coefficient (or the intrinsic diffusion coefficient) under the same conditions. By quantifying the effects of tortuosity factor and porosity on diffusion, we can design and optimize the properties of the material better, which is defined as shown in (S12).

$$D_{eff} = \frac{D_0 \phi}{\tau} \tag{S12}$$

where $D_0$ is the intrinsic diffusion coefficient, i.e., the diffusion coefficient in an ideal unobstructed environment, $\phi$ is the porosity or volume fraction of the conductive phase, which represents the proportion of space available for diffusion and $\tau$ is the tortuosity factor, which represents the ratio of the actual diffusion path to the straight-line distance. The tortuosity factor is equal to or greater than 1, where 1 represents a straight-line path and greater than 1 indicates that the path becomes more tortuous.

**S5: Analysis of three phase characterization parameters of Lithium battery NMC cathode**

As shown in Fig. S3 a-d, the microstructural parameters of the three phases in the NMC materials are exhibited, including the relative diffusion coefficients, phase volume fractions, specific surface areas and tri-phasic boundary characterization parameters. Among them, the gray color represents the standard values of the training data set, the light blue color represents the prediction results of the baseline network, and the green color shows the prediction results of the model proposed in this study. From which it is clearly seen that compared with the baseline network, our model is closer to the actual values of the training data for the four different parameters and thus more accurately reflects the dataset's real characteristics. In the phase volume fraction, the median values generated are highly consistent with the actual values of the samples, which proves the superior performance of the network model. Especially in the specific surface area parameter, the predictions of the white and black phases are close to the actual values, while compared to the Baseline method, which still has a significant bias in the prediction of the three phases. In the TPB density, the predicted values of our model are highly matched with the real values. As for the relative diffusion coefficients, the predictions of the gray and black phases exhibit high similarity to the training samples. In summary, all these characterization parameters show that the algorithmic model proposed in this study exhibits excellent accuracy in the generation of microscopic material quality.

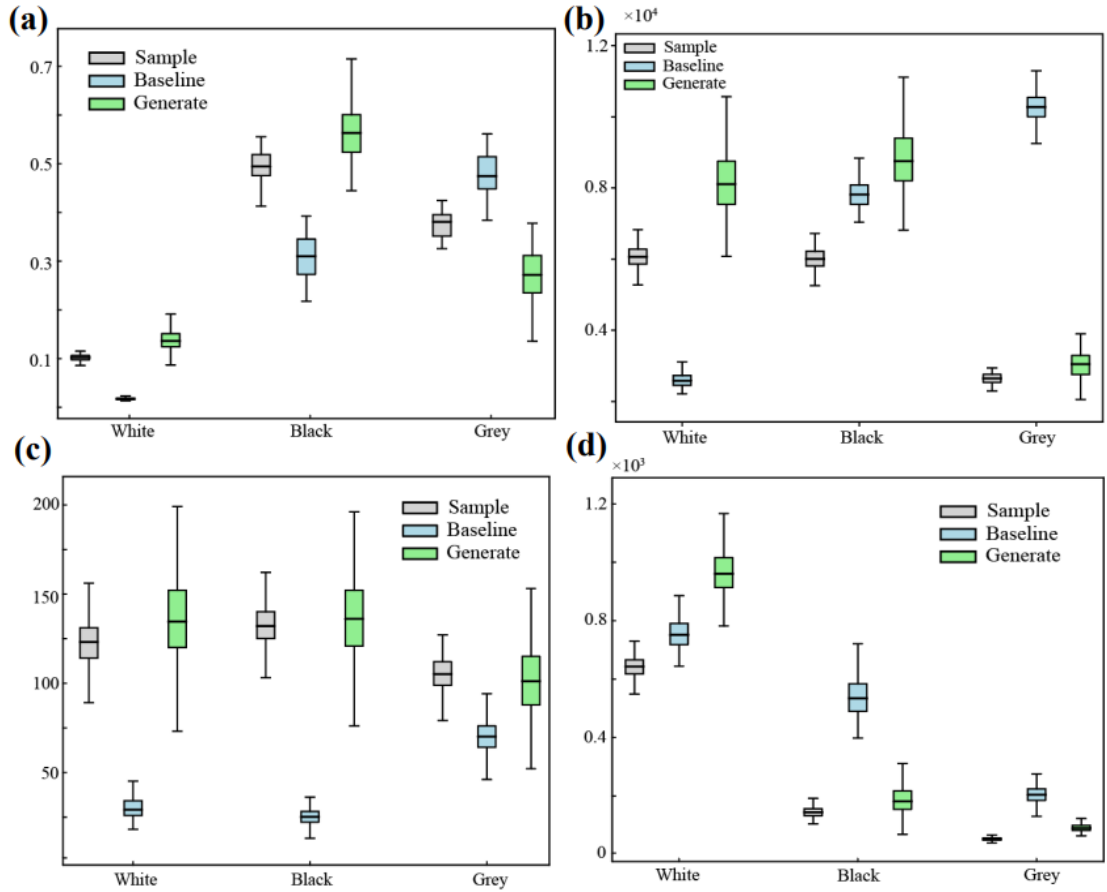

Fig. S3 a-d Comparison of phase volume fractions, specific surface areas, tri-phasic boundaries, and relative diffusion coefficients characterizing parameters, respectively (where the boxes in the a-d plots consist of the minimum, the first quartile $Q_1$ (25% of the data points in the data are below the $Q_1$ value), the median $Q_2$ (indicated by the horizontal line in the box), the third quartile $Q_3$ (25% of the data points in the data are below the $Q_3$ value), and the maximum, with tri-phasic curves of the three models indicated by the different colors and the different types of line segments in the upper-right corners of the plots)

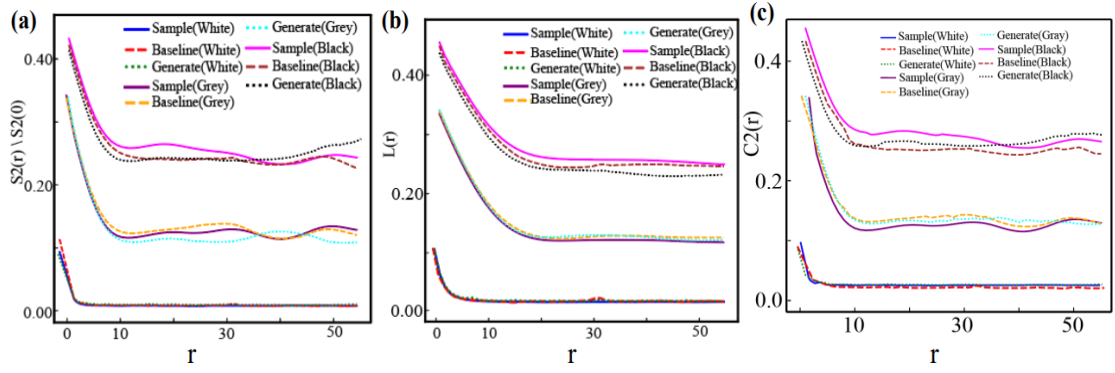

**Fig. S4 a-c** Comparison of the results of the $S_2(r)$, $L(r)$ and $C_2(r)$ respectively for different phases (gray, white, and black)

**S6: Analysis of ablation experiments**

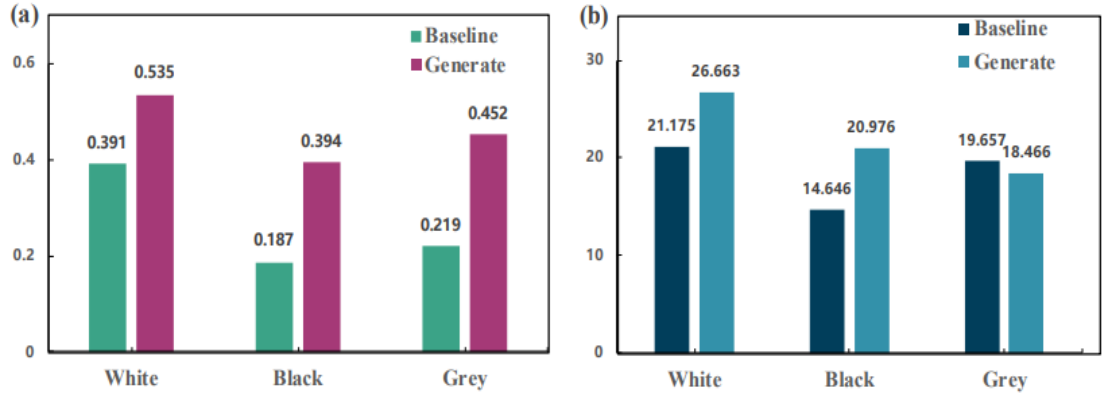

**Fig. S5 a and b** show the comparison of SSIM, PSNR and Baseline for different phases respectively

**Table S8** Impact of different convolutional layers on model generation

| Modual | Multi-scale hierarchical | | | Multi-scale channel aggregation | | | Attention | Metrics | |
|---|---|---|---|---|---|---|---|---|---|
| | MHFA | | | MSCA | | | | | |
| Baseline | Conv 3*3*3 | Conv 5*5*5 | Conv 7*7*7 | Conv 3*3*3 | Conv 5*5*5 | Conv 7*7*7 | CBAM | SSIM | PSNR |
| √ | √ | √ | | | | | √ | 0.278 | 20.200 |
| √ | √ | | √ | | | | √ | 0.313 | 19.977 |
| √ | | √ | √ | | | | √ | 0.286 | 20.899 |
| √ | √ | √ | √ | | | | √ | 0.369 | 20.956 |
| √ | | | | √ | √ | | √ | 0.285 | 18.923 |
| √ | | | | √ | | √ | √ | 0.301 | 19.308 |
| √ | | | | | √ | √ | √ | 0.281 | 19.219 |
| √ | | | | √ | √ | √ | √ | 0.347 | 19.203 |
| √ | √ | √ | √ | √ | √ | √ | √ | **0.461** | **22.035** |

In our algorithm architecture, each component has unique functions and utility. In particular, MHFA significantly enhances the efficiency and stability of the network by exploiting the multi-scale information. In addition, its dense connectivity strategy further accurately captures the scale features and exhibits a high degree of sensitivity to the pixel-level information of the image, which optimizes the reconstruction of the details of the material microstructures as well as its overall integrity. As for the MSCA module, we introduced CAM to focus on processing the inter-channel weighting information, which not only enhances the expressive power of the generation but also improves its versatility. Table

S8 details the impact of each module at each layer in the network, and when we combine these modules together, they significantly enhance the generation of the model.

In order to gain insight into the contribution of the convolutional layers in the MHFA and MSCA modules to the network generation performance, we conducted an empirical study on their necessity and the impact of each convolutional layer on the network generation capability. In the experiments, the network maintains the rest of the modules unchanged and only the convolutional layers in MHFA and MSCA are tuned as shown in Fig. S6. When the experiments are conducted using a two-by-two combination of Conv3×3×3, Conv5×5×5 or Conv7×7×7, the SSIM and PSNR metrics obtained are significantly lower than those obtained using the full three-layer convolutional combination. When all convolutional layers are included in the model, the values of SSIM and PSNR reach 0.461 and 22.035 respectively, indicating that its performance has reached the best. Moreover, as demonstrated in Fig. S7, the generation quality of the slice reconstruction maps of the lithium-ion cathode in terms of microstructure is also significantly improved with the addition of different modules, and the performance the complete model is particularly well.

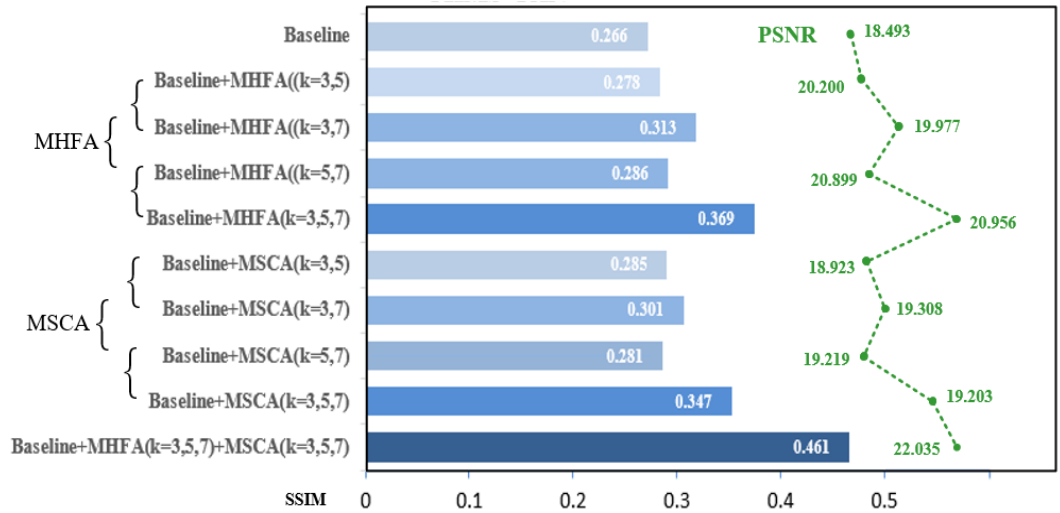

**Fig. S6 Parameter metrics in MHFA, MSCA module using different convolutional layers**

It should be noted that macroscopic image generation tasks often involve the distinction between foreground and background information. There may be certain areas of focus in the model, but unlike this, material microstructure images have their own uniqueness and do not distinguish between foreground and background information. Therefore, the core goal of this task is to reproduce images that are as similar to the original structure as possible. Although SSIM and PSNR are regarded as important indicators to measure image quality in most image generation tasks. However, in this study, we use this indicator focus on the performance of the model as well as its generalizability. Due to a certain degree of randomness in the reconstructed structure, it is difficult to generate a similar structure in the same layer of slices in the real and generated samples. However, in this study, we mainly calculate the SSIM and PSNR for each layer of all the slices of the images in the X, Y and Z planes, which are corresponding to each layer of slices of the real samples, and then we finally seek the average of these slices to make comparisons. Based on which, even though the SSIM and PSNR obtained from the calculation of this paper are relatively lower, we can measure the generative ability of the model itself. We still have to refer to the related indexes of the materials of this field if we want to assess the generating quality of different material fields.

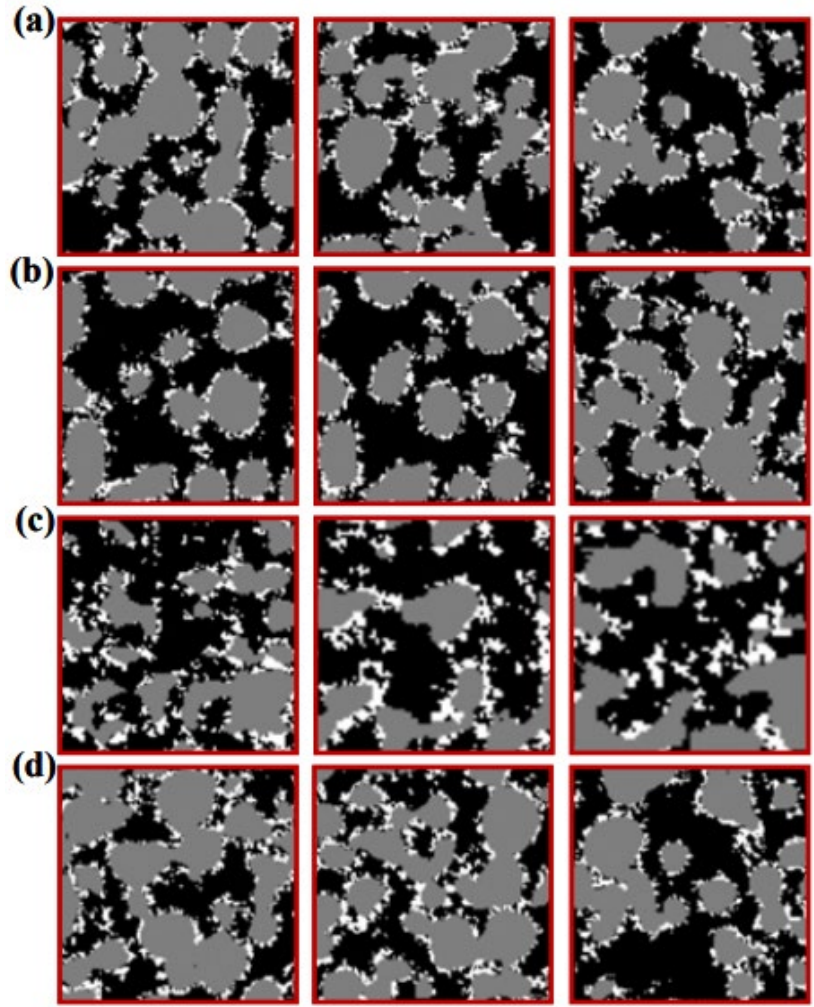

**Fig. S7 Comparison of generated slices under different modules a is the slices in X, Y and Z directions of the 2D structure generated by Baseline model, b is the slices in X, Y and Z directions of the 2D structure generated by Baseline+MHFA+CBAM model, c is the slices in X, Y and Z directions of the 2D structure generated by Baseline+MSCA+CBAM model, and d is the slices in X, Y and Z directions of the 2D structure generated by Baseline+MHFA+MSCA+CBAM model**

**S7: Analysis of characterization parameters of Copper-zinc alloy in X, Y and Z directions**

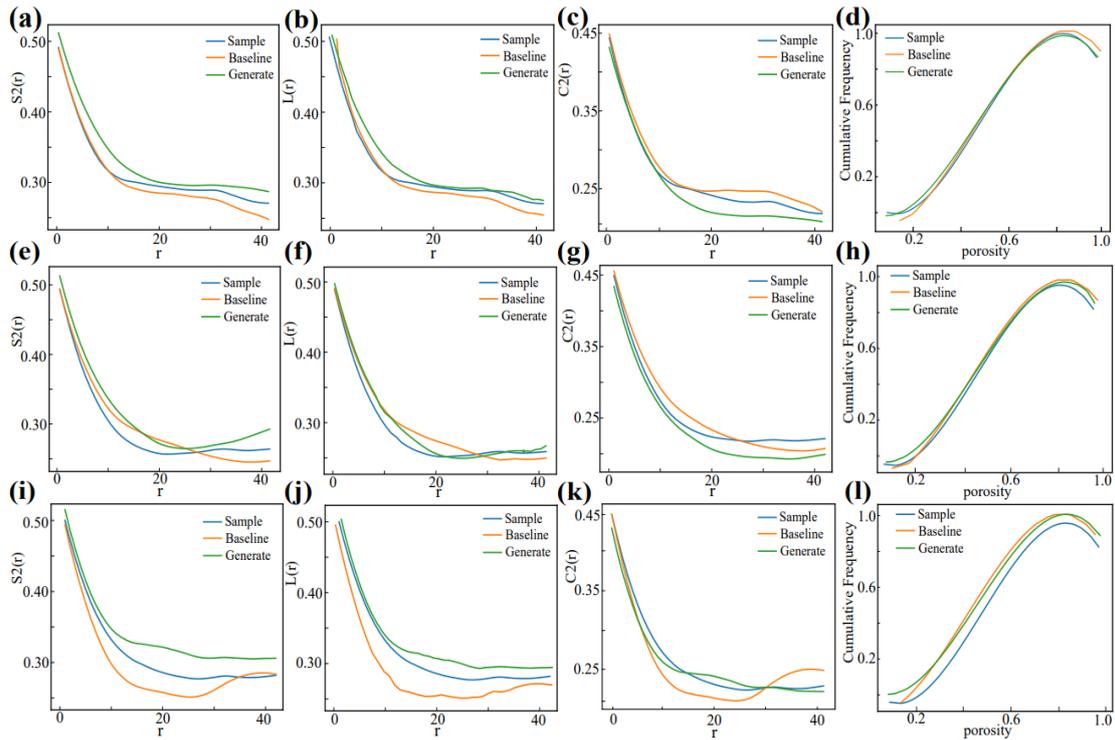

**Fig. S8 a-d, e-h, and i-l represent the comparison of the average results of $S2(r)$, $C2(r)$, $L(r)$ and the cumulative distributions of local porosity in the X, Y and Z directions for Copper-zinc alloy that is clearly anisotropic on the images respectively**

## S8: Reconstructed macroscopic visualization of material samples

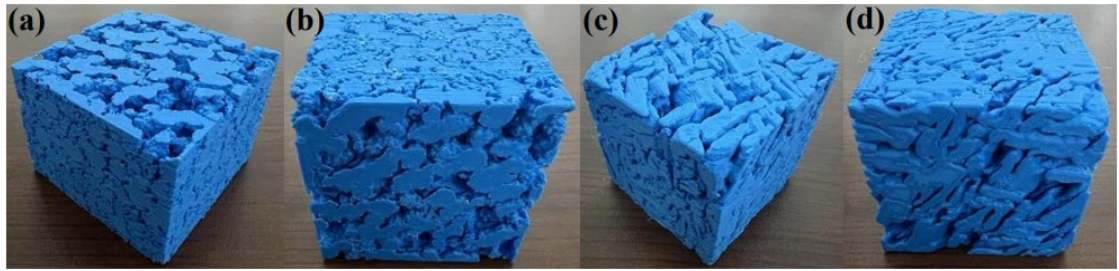

**Fig. S9 a-b shows the 3D printed visualization of Lithium battery NMC cathode reconstructed by the present algorithm, c-d shows the 3D printed visualization of Copper-zinc alloy**

According to the demonstration in Figure S9, an isotropic tri-phasic lithium battery NMC cathode and an anisotropic copper-zinc alloy material were successfully printed. These were accurately reconstructed by the algorithm network using a Creality CR-10 Max model 3D printer. Significantly, the 3D-printed materials exhibit detail richness and visual properties highly consistent with the original materials. This result underscores the efficient capability of the algorithm in structurally reconstructing complex materials and highlights the potential of the method for important applications in macro-analysis and ensuring model manufacturability, thus offering an effective new avenue for in-depth analysis of material physical properties.